\documentclass[10pt,twocolumn,letterpaper]{article}
\usepackage{enumitem}
\usepackage{cvm}
\usepackage{times}
\usepackage{epsfig}
\usepackage{graphicx}
\usepackage{amsmath}
\usepackage{amssymb}
\usepackage{stmaryrd}
\usepackage{booktabs} 
\usepackage{multirow}
\usepackage[normalem]{ulem}
\usepackage{booktabs}
\usepackage{color}
\definecolor{amber}{rgb}{1.0, 0.75, 0.0}
\definecolor{bittersweet}{rgb}{1.0, 0.44, 0.37}

\usepackage[pagebackref=true,breaklinks=true,letterpaper=true,colorlinks,bookmarks=false]{hyperref}

\newcommand{\keywords}[1]{{\bf \emph{Keywords: #1}}}

\cvmfinalcopy % *** Uncomment this line for the camera-ready version

\def\cvmPaperID{257} % *** Enter the cvm Paper ID here

\ifcvmfinal\fi
\begin{document}

%%%%%%%%% TITLE
\title{OAAFormer: Robust and Efficient Point Cloud Registration Through Overlapping-Aware Attention in Transformer}

\author{Junjie Gao\\
Shandong University\\
% Institution1 address\\
{\tt\small junjie.gao95.m@gmail.com}
% For a paper whose authors are all at the same institution,
% omit the following lines up until the closing ``}''.
% Additional authors and addresses can be added with ``\and'',
% just like the second author.
% To save space, use either the email address or home page, not both
\and
Qiujie Dong\\
Shandong University\\
% First line of institution2 address\\
{\tt\small qiujie.jay.dong@gmail.com}
\and
Ruian Wang\\
Shandong University\\
% First line of institution2 address\\
{\tt\small wra.time@gmail.com}
\and
Shuangmin Chen\\
Qingdao University of Science and Technology\\
% First line of institution2 address\\
{\tt\small csmqq@163.com}
\and
Shiqing Xin*\\
Shandong University\\
% First line of institution2 address\\
{\tt\small xinshiqing@sdu.edu.cn}
\and
Changhe Tu\\
Shandong University\\
% First line of institution2 address\\
{\tt\small chtu@sdu.edu.cn}
\and
Wenping Wang\\
Texas A\&M University\\
% First line of institution2 address\\
{\tt\small wenping@tamu.edu}
}

% {\small\url{http://www.author.org/~second}}
% }

\maketitle
% \thispagestyle{empty}
%%%%%%%%% ABSTRACT
\begin{abstract}
In the domain of point cloud registration, the coarse-to-fine feature matching paradigm has received substantial attention owing to its impressive performance. This paradigm involves a two-step process: first, the extraction of multi-level features, and subsequently, the propagation of correspondences from coarse to fine levels. Nonetheless, this paradigm exhibits two notable limitations.
Firstly, 
the utilization of the Dual Softmax operation has the potential to promote one-to-one correspondences between superpoints, inadvertently excluding valuable correspondences. This propensity arises from the fact that a source superpoint typically maintains associations with multiple target superpoints.
Secondly, it is imperative to closely examine the overlapping areas between point clouds, as only correspondences within these regions decisively determine the actual transformation. Based on these considerations, we propose {\em OAAFormer} to enhance correspondence quality.
On one hand, we introduce a soft matching mechanism, facilitating the propagation of potentially valuable correspondences from coarse to fine levels. 
Additionally, we integrate an overlapping region detection module to minimize mismatches to the greatest extent possible. Furthermore, we introduce a region-wise attention module with linear complexity during the fine-level matching phase, designed to enhance the discriminative capabilities of the extracted features.
Tests on the challenging 3DLoMatch benchmark demonstrate that our approach leads to a substantial increase of about 7\% in the inlier ratio, as well as an enhancement of 2-4\% in registration recall.
Finally, to accelerate the prediction process, we replace the conventional RANSAC algorithm with the selection of a limited yet representative set of high-confidence correspondences.
Remarkably, this change results in a 100x speedup while still maintaining comparable registration performance.

% All experimental results on three public datasets validate the effectiveness of the pro- posed method. For example, tests on the challenging 3DLoMatch benchmark, OAAFormer improves the inlier ratio by over 7\%, and boosts registration recall by over 2\%.

\end{abstract}


%%%%%%%%% BODY TEXT
\section{Introduction}
The task of point cloud registration involves determining a rigid transformation that aligns one point cloud with another. This challenge is of fundamental importance in the fields of computer vision and robotics and has wide-ranging applications, including 3D reconstruction~\cite{azinovic2022neural,deng2022depth,li2022bnv,wang2021neural}, SLAM~\cite{Barros2022ACS,Chaplot2020LearningTE,MurArtal2015ORBSLAMAV,Teed2021DROIDSLAMDV}, and autonomous driving~\cite{Chitta2021NEATNA,Hu2022STP3EV,Prakash2021MultiModalFT,Yang2020SurfelGANSR}. A common approach to this task involves two key stages: point feature matching and globally consistent refinement.
During the point feature matching phase, the goal is to generate a set of initial correspondences with a high inlier ratio, ideally including as many true correspondences as possible while minimizing false ones. However, achieving this objective is a formidable challenge due to inherent noise and disparities in the input point clouds, as well as the possibility of partial overlap between them.
Conversely, in the globally consistent refinement step, the focus shifts to rapidly identifying a subset of correspondences capable of consistently encoding the actual transformation through further refinement.

\begin{figure}[t]
  \centering
  \includegraphics[width=\linewidth]{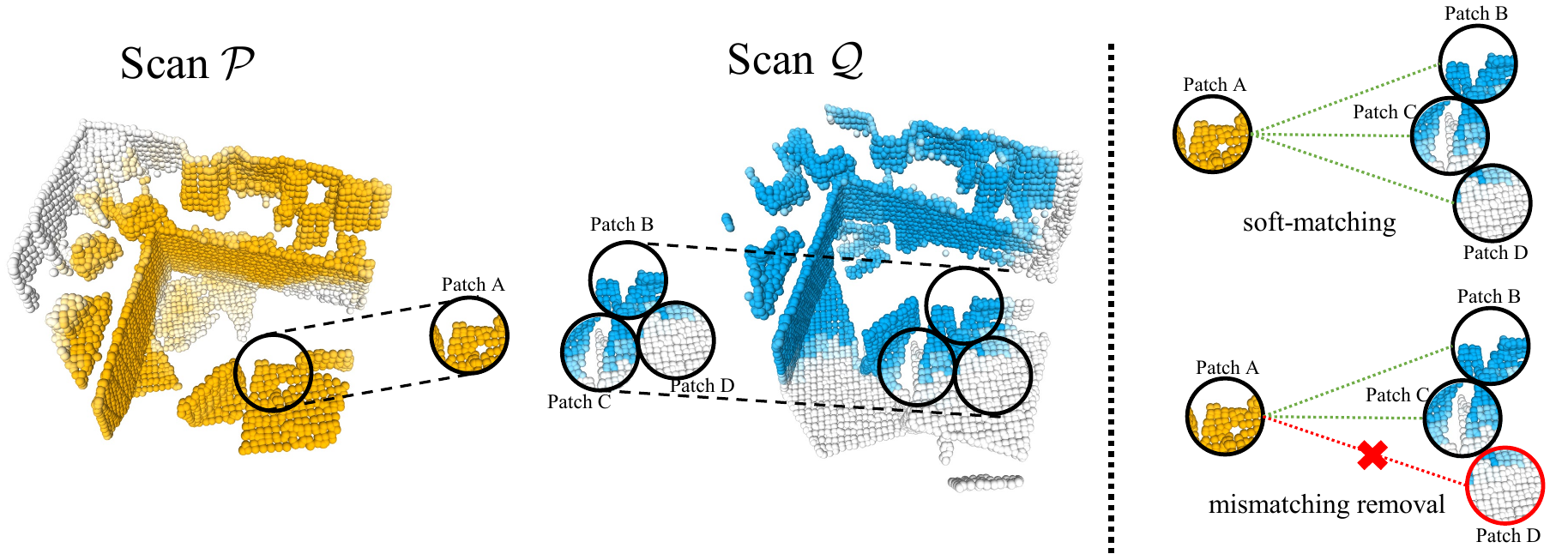}
  \vspace{-15pt}
  \caption{
  Considering that in coarse-level matching, the correspondence between a source superpoint and a target superpoint inherently embodies a patch-based mapping, there exists the possibility of overlooking potentially valuable correspondences due to the use of the Dual Softmax operation~\cite{cheng2021improving, Qin2022GeometricTF}.
To ameliorate this concern, we introduce a soft matching mechanism that permits one-to-many correspondences, effectively addressing this limitation. Moreover, our network incorporates a dedicated module for predicting overlap regions, which serves the purpose of filtering out significantly unhelpful correspondences. It is noted that the intensity of the color (yellow or blue) indicates the overlapping score.
  }
  \label{fig.1}
\end{figure}

% \SQ{to revise}
% The use of feature detectors reduces the matching search space and achieves considerable performance on real large-scale benchmarks. 
% poses a challenge to repeatability, {\itshape i.e.}, down-sampling increases the risk where a specific point loses its corresponding point in the other frame, which constrains the performance of detector-based methods. Recently, the coarse-to-fine matching mechanism~\cite{Qin2022GeometricTF,Yu2021CoFiNetRC}, being different from the detector-based methods, has drawn considerable attention due to its high performance. The process commences with the downsampling of the input point cloud into superpoints. By employing a superpoint as a representative for a point patch, dense correspondences can be effectively established. 
While a substantial body of literature~\cite{Yew20183DFeatNetWS,Li2019USIPUS,Bai2020D3FeatJL,Wu2020SKNetDL,Huang2020PREDATORRO} has focused on the extraction of discriminative features to enhance correspondence quality, the inherent sparsity and disparities in point clouds, along with potential partial overlap, present persistent challenges. 
Recently, the coarse-to-fine matching paradigm~\cite{Qin2022GeometricTF,Yu2021CoFiNetRC} has garnered significant attention for its impressive performance. This paradigm begins by downsampling the input point cloud into superpoints and establishing correspondences between these superpoints, where each superpoint inherently represents a point patch. Subsequently, sparse correspondences are propagated to encompass more points, resulting in the generation of dense correspondences.

However, accurately matching a superpoint from one scan to another can be challenging, as the corresponding point patches may not exhibit perfect alignment. As illustrated in Fig.~\ref{fig.1}, suppose we have two input point clouds $\mathcal{P}$ and $\mathcal{Q}$. The superpoint $A$ is associated with $B,C,D$ simultaneously. Yet, the use of the Dual Softmax operation~\cite{cheng2021improving, Qin2022GeometricTF} within the coarse-to-fine paradigm has the potential to enforce one-to-one correspondences between superpoints, unintentionally excluding valuable correspondences. This represents the first limitation of the coarse-to-fine paradigm.
On the other hand, it is crucial to examine the overlapping regions between point clouds, as only correspondences within these areas decisively determine the actual transformation. Consequently, there is a pressing need to enhance the discriminability of the features extracted from points within these overlapping regions to improve the overall performance of the coarse-to-fine paradigm.

Motivated by these considerations, we propose a robust matching network, named {\em OAAFormer}, with the explicit objective of augmenting the performance of the coarse-to-fine matching paradigm. This augmentation is achieved through the systematic integration of a suite of strategies meticulously designed to elevate the quality of correspondences.
Firstly, OAAFormer employs a sophisticated soft matching mechanism, with the explicit purpose of seamlessly propagating potentially valuable correspondences from the coarse to the fine levels of the matching process.
Secondly, OAAFormer incorporates an intricately designed overlapping region detection module, strategically engineered to minimize the probability of mismatches.
Thirdly, it introduces a region-wise attention module characterized by linear computational complexity, meticulously designed to enhance the discriminative capabilities of the extracted features during the fine-level matching phase.
Empirical validation underscores the efficacy of these strategies.
For instance, tests on the exacting 3DLoMatch benchmark show that our approach yields a substantial increase of approximately 7\% in the inlier ratio, as well as a discernible enhancement of 2-4\% in registration recall. 
Furthermore, we replace the conventional RANSAC algorithm~\cite{Fischler1981RandomSC} with the selection of a limited yet representative set of high-confidence correspondences for accelerating the prediction process.

In summary, the main contributions of this work are as follows:
\begin{itemize}
    \item We use a soft matching mechanism to facilitate the propagation of potentially valuable correspondences from coarse to fine levels, 
    which finally results in a substantial increase in the inlier ratio and registration recall.
    \item We introduce a region-wise attention module with linear complexity during the fine-level matching phase, designed to enhance the discriminative capabilities of the extracted features.    
    \item Through the replacement of the inefficient RANSAC algorithm with a more intelligent mechanism for selecting high-confidence correspondences, we achieve a remarkable 100x acceleration in the prediction process.
\end{itemize}

\section{Related work}
\subsection{Point cloud registration}
The construction of feature descriptors with specific characteristics proves to be an effective means of encoding the curvature of the underlying surface, providing valuable information for the alignment of point clouds. In previous research, a multitude of traditional methods~\cite{Johnson1999UsingSI,Tombari2010UniqueSC,Tombari2010UniqueSO,Rusu2009FastPF,Guo20133DFF} have relied on handcrafted features to craft such descriptors. With the proliferation of deep learning techniques, various learning-based descriptors~\cite{Zeng20163DMatchLL,Gojcic2018ThePM,Choy2019FullyCG,Ao2020SpinNetLA,Thomas2019KPConvFA,Wang2021YouOH,Wang2023RoRegPP,Yu2022RIGARA} have been introduced to enhance the expressiveness of these feature descriptors.
However, the task of identifying valuable correspondences between points based solely on geometric descriptors remains a challenging one, primarily due to the presence of various defects in the input point clouds, including noise, disparities, and partial overlapping. Consequently, approaches such as the Random Sample Consensus (RANSAC) algorithm~\cite{Fischler1981RandomSC, Myatt2002NAPSACHN, Barth2017GraphCutR} or meticulously designed neural networks~\cite{Pais20193DRegNetAD,Choy2020DeepGR,Bai2021PointDSCRP,Chen2022SC2PCRAS} are frequently employed to address this challenge. These methods aim to eliminate mismatches, even when dealing with points possessing similar features, ultimately resulting in a more robust and accurate registration outcome.

Additionally, a variety of keypoint detectors tailored for rigid registration tasks have emerged. For instance, D3Feat~\cite{Bai2020D3FeatJL} 
introduces
a keypoint selection strategy that overcomes the inherent density variations of 3D point clouds. However, this approach does not fully account for overlapping areas and exhibits reduced robustness in scenarios with low overlap.
Another noteworthy method, Predator~\cite{Huang2020PREDATORRO},
develops an overlap-attention block for early information exchange between the latent encodings of the two point clouds.
Keypoints are selected based on both saliency and overlap scores. While Predator~\cite{Huang2020PREDATORRO} demonstrates substantial improvements over existing methods across indoor and outdoor benchmarks, challenges persist in extracting a set of representative keypoints.

Recently, the coarse-to-fine paradigm has garnered attention for enhancing the quality of correspondences, not only in 2D image matching~\cite{Li2020DualResolutionCN,Zhou2020Patch2PixEP,Sun2021LoFTRDL,Huang2022AdaptiveAF} but also in the domain of point cloud registration~\cite{Yu2021CoFiNetRC,Qin2022GeometricTF}. For instance, CofiNet~\cite{Yu2021CoFiNetRC} incorporates an optimal transport~\cite{cuturi2013sinkhorn, peyre2019computational} matching layer to establish correspondences between mutually nearest patches and subsequently refines these correspondences at the fine-level stage. In a similar vein, Geotransformer~\cite{Qin2022GeometricTF} introduces a self-attention mechanism to learn geometric features, thereby improving the matching accuracy between superpoints based on whether their neighboring patches overlap.

In this paper, we further enhance the coarse-to-fine mechanism through a set of strategies, including (1) a soft matching mechanism that streamlines the propagation of potentially valuable correspondences from coarse to fine levels and (2) a region-wise attention module characterized by linear complexity during the fine-level matching phase. 

% Recently, the coarse-to-fine paradigm 
% has been used to improve the quality of correspondences,
% not only in 2D image matching~\cite{Li2020DualResolutionCN,Zhou2020Patch2PixEP,Sun2021LoFTRDL,Huang2022AdaptiveAF} but also in point cloud registration~\cite{Yu2021CoFiNetRC,Qin2022GeometricTF}.
% For example, CofiNet~\cite{Yu2021CoFiNetRC} employs an optimal transport matching layer to match mutual-nearest patches, and then generates refined correspondences at the fine-level stage.
%  Given that superpoints are matched based on whether their neighboring patches overlap, Geotransformer~\cite{Qin2022GeometricTF} proposes 
%  a self-attention mechanism
%  to learn geometric feature for improving the matching accuracy between superpoints.
%  In this paper, 
%  we further improve the coarse-to-fine mechanism by a set of strategies,
%  including (1)~a soft matching mechanism to facilitate the
% propagation of potentially valuable correspondences
% from coarse to fine levels and (2)~a region-wise attention module with linear complexity during the fine-level matching phase to enhance the discriminative capabilities of
% the extracted features.

\subsection{Efficient Transformer}
% In the standard Transformer model~\cite{Vaswani2017AttentionIA}, the memory cost experiences a quadratic increase due to matrix multiplication, which has become a bottleneck when dealing with long sequences. Recently, various efficient Transformer variants have been introduced~\cite{Katharopoulos2020TransformersAR, Shen2018EfficientAA,Wang2020LinformerSW,Zaheer2020BigBT,Wu2021FastformerAA}. For instance, the Linear Transformer~\cite{Katharopoulos2020TransformersAR} formulates self-attention as a linear dot product of kernel feature maps and leverages the associativity property of matrix products to reduce computational complexity. BigBird~\cite{Zaheer2020BigBT} combines local and global attention mechanisms at specific positions and incorporates random attention for selected token pairs. FastFormer~\cite{Wu2021FastformerAA} employs an additive attention mechanism to model global contexts, achieving effective context modeling with linear complexity.
In the standard Transformer model~\cite{Vaswani2017AttentionIA}, the memory cost exhibits a quadratic increase due to matrix multiplication, which has become a bottleneck when handling long sequences. Recently, several efficient Transformer variants have been introduced~\cite{Katharopoulos2020TransformersAR, Shen2018EfficientAA, Wang2020LinformerSW, Zaheer2020BigBT, Wu2021FastformerAA}. For example, the Linear Transformer~\cite{Katharopoulos2020TransformersAR} reformulates self-attention as a linear dot product of kernel feature maps and exploits the associativity property of matrix products to reduce computational complexity. BigBird~\cite{Zaheer2020BigBT} combines local and global attention mechanisms at specific positions and introduces random attention for selected token pairs. FastFormer~\cite{Wu2021FastformerAA} employs an additive attention mechanism to model global contexts, achieving effective context modeling with linear complexity.
Inspired by these advancements, we propose a region-wise attention module with linear complexity during the fine-level matching phase, meticulously designed to enhance the discriminative capabilities of the extracted features for points within overlapping areas.
% In the vanilla Transformer~\cite{Vaswani2017AttentionIA}, the memory cost
% climbs in the second order due to matrix multiplication, which has become a bottleneck in Transformer when dealing with long sequences. Recently, some efficient transformer works have been proposed~\cite{Katharopoulos2020TransformersAR, Shen2018EfficientAA,Wang2020LinformerSW,Zaheer2020BigBT,Wu2021FastformerAA}, such as Linear Transformer~\cite{Katharopoulos2020TransformersAR} expresses self-attention as a linear dot product of kernel feature maps and makes use of the associativity property of matrix products to reduce the computational complexity. BigBird~\cite{Zaheer2020BigBT} combines local attention and global attention at certain positions and utilizes random attention on several randomly selected token pairs. FastFormer~\cite{Wu2021FastformerAA} uses an additive attention mechanism to model global contexts, achieving effective context modeling with linear complexity. 
% Inspired by these works, we replace the conventional RANSAC algorithm with the selection of a limited yet representative set of high-confidence correspondences for accelerating the prediction process while still maintaining comparable registration performance.

\begin{figure*}[t]
  \centering
  \includegraphics[width=\linewidth]{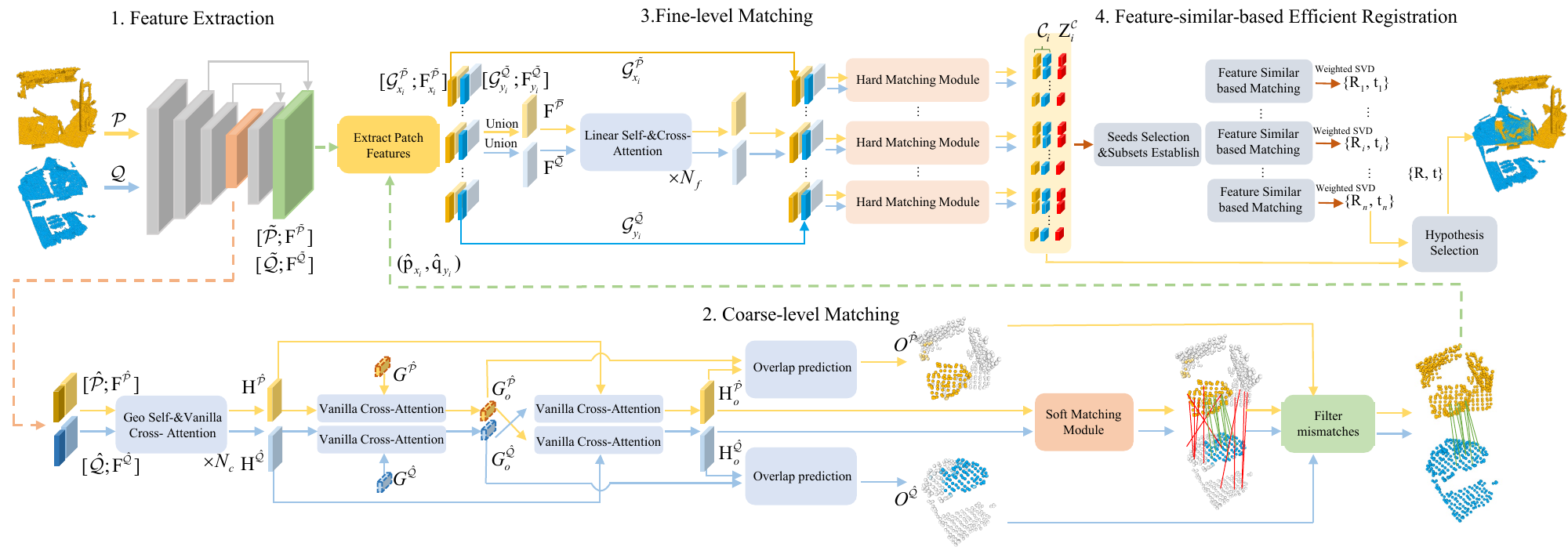}
  \vspace{-12pt}
  \caption{The backbone network down-samples the input point cloud to extract features in multiple resolutions. In the coarse-level matching step, a soft matching mechanism is employed to establish one-to-many correspondences between superpoints, while the overlapping detection module is introduced to eliminate mismatches outside the overlapping region. In the fine-level matching step, the correspondences between superpoints propagate to the dense point sets $\tilde{\mathcal{P}}$ and $\tilde{\mathcal{Q}}$, and the matching capability of features is enhanced through linear attention modules. Ultimately, the transformation estimation is calculated using an efficient estimation module based on feature similarity.}
  \label{fig.2}
\end{figure*}

%%%%%%%%%%%%%%%%%%%%%%%%%%%%%%%%%%%%%%%%%%%%%%%%%%%%%%%%%%%%%%%%%%%%%%%%%%%%%%%%%%%%%%%%%%%%%%%%%%%%%%%
\section{Method}
\subsection{Pipeline}
Suppose that we have a source point cloud $\mathcal{P}=\left\{\mathbf{p}_{i} \in \mathbb{R}^{3} \mid i=1, \ldots, N\right\}$ and a target point cloud $\mathcal{Q}=\left\{\mathbf{q}_{i} \in \mathbb{R}^{3}\mid i=1,\ldots, M\right\}$.
% We define point clouds $\mathcal{P}=\left\{\mathbf{p}_{i} \in \mathbb{R}^{3} \mid i=1, \ldots, N\right\}$ and $\mathcal{Q}=\left\{\mathbf{q}_{i} \in \mathbb{R}^{3}\mid i=1,\ldots, M\right\}$, as source and target, respectively. 
The objective of rigid registration is to estimate the unknown rigid transformation $\mathbf{T}=\left\{\mathbf{R},\ \mathbf{t}\right\}$, where $\mathbf{R} \in \text{SO}(3)$ represents a rotation matrix and $\mathbf{t} \in \mathbb{R}^{3}$ represents a translation vector. 
% The goal of rigid registration is to recover the unknown rigid transformation $\mathbf{T}=\left\{\mathbf{R},\ \mathbf{t}\right\}$, where $\mathbf{R} \in \text{SO}(3)$ is a rotation matrix and $\mathbf{t} \in \mathbb{R}^{3}$ is a translation vector. 
Let $$\mathcal{C}^*=\{\mathbf{p}_{i_k}\mapsto\mathbf{q}_{j_k},k=1,2,\cdots,K\}$$ denote the set of ground-truth correspondences between~$\mathcal{P}$ and~$\mathcal{Q}$.
The true transformation~$\mathbf{T}$ should accurately map each $\mathbf{p}_{i_k}\in \mathcal{P}$ to $\mathbf{q}_{j_k}\in \mathcal{Q}$,
meaning that it should minimize the difference vector $$\mathbf{R} \mathbf{p}_{i_k}+\mathbf{t}-\mathbf{q}_{j_k}$$ to be nearly zero.
In real-world scenarios, where $\mathcal{C}^*$ remains elusive, the prevailing approach involves extracting a subset of correspondences considered reasonably reliable between two point clouds. Subsequently, the estimation of the transformation matrix relies on correspondence consistency.

% The transformation can be solved by:
% \begin{equation}
% \min_{\mathbf{R,t}} {\sum}_{{\mathbf{(\mathbf{p}_{x_i}^*, \mathbf{q}_{y_i}^*)}} \in \mathcal{C}^*} \|\mathbf{R} \cdot \mathbf{p}_{x_i}^*+\mathbf{t}-\mathbf{q}_{y_i}^*\|_2^2
% \end{equation}
% Here $\mathcal{C}^*$ is the set of ground-truth correspondences between $\mathcal{P}$ and $\mathcal{Q}$. Since $\mathcal{C}^*$ is unknown in reality, we need to first build a set of putative correspondences between two point clouds and then estimate the alignment transformation.

As shown in  Fig.~\ref{fig.2},
our algorithmic pipeline includes the following stages:
% \begin{itemize}[noitemsep,topsep=0pt,parsep=0pt,partopsep=0pt]
%     \item[S1.] Downsample the point cloud and extract the point-wise features simultaneously, like that achieved in KPConv~\cite{Thomas2019KPConvFA}. \SQ{simultaneously?}
%     \item[S2.] \SQ{?}
%     \item[S3.] \SQ{?}
% \end{itemize}

% 1. For feature extraction, KPConv serves as the backbone to downsample the point cloud and extract multi-level features. Where the first and last level points with their associated features are used for subsequent matching.
% 2. For coarse level matching, geotransformer is first used to improve the global modeling capability of superpoint's features, and then the overlap region estimation is realized through the proposed overlap region detection module. After that, a soft matching strategy that allows one-to-many matching is applied to get the patch-
% level correspondences, while the previously estimated over-lap regions are used to filter the mismatches.
% 3. For fine-level matching,  a region-wise attention module with linear complexity is introduced to enhance the discriminant ability of the extracted features.
% 4. 

\begin{itemize}[noitemsep,topsep=0pt,parsep=0pt,partopsep=0pt]
    \item[S1.] During the feature extraction stage, we employ KPConv~\cite{Thomas2019KPConvFA} as the backbone to downsample the point cloud and extract multi-level features. Subsequently, we select sample points from both the first and last levels for the subsequent matching process.
    \item[S2.] In the coarse-level matching stage, we employ the Geotransformer~\cite{Qin2022GeometricTF} to produce the geometric features of the superpoints. Additionally, we estimate the overlap region using a dedicated detection module specifically designed for this purpose. See Sec~\ref{sec.3.1}.    
    \item[S3.] Subsequently, we introduce a soft matching mechanism to extract valuable correspondences at the patch level, followed by a filtering step to remove potential mismatches. See Sec~\ref{sec.3.1}. 
    \item[S4.] In the fine-level matching stage, we introduce a region-wise attention module characterized by linear complexity. This module is designed to enhance the discriminative capabilities of the extracted features. See Sec~\ref{sec.3.2}. 
    \item[S5.] In the pose estimation stage, we have devised an efficient seeding mechanism for the identification of high-confidence correspondences, aiming to expedite the process. See Sec~\ref{sec.3.3}. 
\end{itemize}

\subsection{Coarse-level Matching}\label{sec.3.1}
\noindent {\bfseries Intra- and Inter- Consistency Enhancement:} 
In the coarse-level matching phase, considering superpoints $\hat{\mathcal{P}}$ and $\hat{\mathcal{Q}}$ with associated features $\mathbf{F}^{\hat{\mathcal{P}}} \in \mathbb{R}^{|\hat{\mathcal{P}}|\times d_t}$ and $\mathbf{F}^{\hat{\mathcal{Q}}}\in \mathbb{R}^{|\hat{\mathcal{Q}}|\times d_t}$, we alternately apply the self-attention layer within each point cloud and the cross-attention layer between point clouds $\mathit{N}_{c}$ times to enhance the consistency. It's worth noting that we utilize the geometry-aware self-attention mechanism~\cite{Qin2022GeometricTF} instead of the standard self-attention~\cite{Vaswani2017AttentionIA}, as the former is better suited for capturing long-range contextual information.

%In Coarse-level matching, given superpoints $\hat{\mathcal{P}}$ and $\hat{\mathcal{Q}}$ with 
%  associated features ${\mathbf{F}}^{\hat{\mathcal{P}}} \in {\mathbb{R}^{|\hat{\mathcal{P}}|\times {d}}}$ and $\mathbf{F}^{\hat{\mathcal{Q}}} \in {\mathbb{R}^{|\hat{\mathcal{Q}}|\times {d}}}$, we swap the self-attention layer and cross-attention layer repeatedly for $\mathit{N}_{c}$ times to improve the consistency of intra- and inter-point cloud. Here, we use the geometry-aware self-attention~\cite{Qin2022GeometricTF} instead of the vanilla version~\cite{Vaswani2017AttentionIA}, as the former can better capture long-distance contexts.

% 采用注意力模块的本质上是建立一种全局相关性，然而，与大范围非重叠超点建立相关性会带来歧义。
% 对于点云配准任务，建立重叠点之间的相关性是保证几何一致性的关键，因此我们引入一个token-based 注意力机制来提升重叠和非重叠区域的区分能力。
\noindent {\bfseries Overlap Region Detection Module:} 
% \SQ{to revise}
% The essence of using the attention module is to establish a global correlation, however, establishing a correlation with a large range of non-overlapping points leads to ambiguity. For point cloud registration task, establishing the correlation between overlapping points is the key to ensure geometric consistency, so we introduce a token-based attention mechanism to improve the discrimination between overlapping and non-overlapping regions. Specifically, we employ a token $G^{\hat{\mathcal{P}}} \in {\mathbb{R}^{ 1 \times {d}}}$ for embedding the information pertaining to the overlapping region. The initialization of $G^{\hat{\mathcal{P}}}$ is achieved by applying a max-pooling operation on the enhanced feature $\mathbf{H}^{\hat{\mathcal{P}}}$. And then, a cross-attention operation is employed to update token $G^{\hat{\mathcal{P}}}$ to obtain $G^{\hat{\mathcal{P}}}_o$ , enabling it to discriminate between overlapping and non-overlapping regions. Specifically, the query arise from the initialized token $G^{\hat{\mathcal{P}}}$, while keys and values arise from feature $\mathbf{H}^{\hat{\mathcal{P}}}$. 
% \SQ{in the following subsection, please pay attention to the symbols}
To enhance the distinction between overlapping and non-overlapping regions, we introduce a token-based attention mechanism. Specifically, we employ a token, denoted as $G^{\hat{\mathcal{P}}}$, to encapsulate information related to the overlapping region. The initialization of $G^{\hat{\mathcal{P}}}$ is accomplished through a max-pooling operation applied to the augmented feature set $\mathbf{H}^{\hat{\mathcal{P}}}$. Subsequently, we employ a cross-attention operation to update the token $G^{\hat{\mathcal{P}}}$, resulting in $G^{\hat{\mathcal{P}}}_o$. This updated token is instrumental in distinguishing between overlapping and non-overlapping regions. In the implementation, the query originates from the initialized token $G^{\hat{\mathcal{P}}}$, while both keys and values are derived from the feature set $\mathbf{H}^{\hat{\mathcal{P}}}$.
Finally, the tokens obtained, namely $G^{\hat{\mathcal{P}}}_o$ and $G^{\hat{\mathcal{Q}}}_o$, serve as guiding elements for updating the original features $\mathbf{H}^{\hat{\mathcal{P}}}$ and $\mathbf{H}^{\hat{\mathcal{Q}}}$ through an additional cross-attention operation. This is formally represented as:

% This approach is motivated by the following rationale: The fundamental purpose of the attention module is to establish global correlations, but establishing correlations with a broad range of non-overlapping points introduces ambiguity. In the context of point cloud registration, however, the establishment of correlations between overlapping points is pivotal for ensuring geometric consistency. Hence, we introduce a token-based attention mechanism to enhance the differentiation between overlapping and non-overlapping regions.
\begin{equation}
    G^{\hat{\mathcal{P}}}_o = \operatorname{vanillaTransformer}(Q = G^{\hat{\mathcal{P}}}, K=V=\mathbf{H}^{\hat{\mathcal{P}}}),
\end{equation}
Here, $G^{\hat{\mathcal{Q}}}_o$ is computed in the same way.

Subsequently, the obtained tokens $G^{\hat{\mathcal{P}}}_o$ and $G^{\hat{\mathcal{Q}}}_o$ are used as guide items to update the original features $\mathbf{H}^{\hat{\mathcal{P}}}$ and $\mathbf{H}^{\hat{\mathcal{Q}}}$ through another cross-attention operation:
\begin{equation}
    \mathbf{H}^{\hat{\mathcal{P}}}_o = \operatorname{vanillaTransformer}(Q = \mathbf{H}^{\hat{\mathcal{P}}}, K=V=G^{\hat{\mathcal{Q}}}_o),
\end{equation}
Here, $\mathbf{H}^{\hat{\mathcal{Q}}}_o$ is computed in the same way. During this process, $\mathbf{H}^{\hat{\mathcal{P}}}$ and 
$\mathbf{H}^{\hat{\mathcal{Q}}}$ are updated to  $\mathbf{H}_\text{o}^{\hat{\mathcal{P}}}$ and 
$\mathbf{H}_\text{o}^{\hat{\mathcal{Q}}}$, respectively, such that they are aware of the overlapping region between $\hat{\mathcal{P}}$
and $\hat{\mathcal{Q}}$. The overlapping-aware mechanism is highly advantageous as it enhances the ability to effectively discriminate between the overlapping region and the non-overlapping region.

To further identify the location of the overlapping region, we have devised an additional module designed to assign a probability score indicating the likelihood that a point is situated within the overlap region.
Specifically, we project the decoded tokens~$G_\text{o}^{\hat{\mathcal{P}}}$ and~$G_\text{o}^{\hat{\mathcal{Q}}}$ through matrix multiplication and the sigmoid function to create the weight mapping. The weight map~$w^{\hat{\mathcal{P}}}$ is employed to enhance the overlap information within the features. Subsequently, a linear projection operator~$\mathbf{W}^O \in \mathbb{R}^{d_t \times 1}$, and a a sigmoid function are applied to obtain the overlapping confidence:
% with dimensions matching that of the token, along with a sigmoid function, are applied to obtain the overlapping confidence:
\begin{equation}
w^{\hat{\mathcal{P}}} = \operatorname{sigmoid}((\mathbf{H}_\text{o}^{\hat{\mathcal{P}}})^T G_\text{o}^{\hat{\mathcal{P}}})   
\end{equation}
\begin{equation}
O^{\hat{\mathcal{P}}} = \operatorname{sigmoid}((w^{\hat{\mathcal{P}}}\odot\mathbf{H}^{\hat{\mathcal{P}}}_\text{o}+\mathbf{H}^{\hat{\mathcal{P}}}_\text{o})\mathbf{W}^O)
\end{equation}
% \SQ{what is $\theta^{\hat{\mathcal{P}}}$?}
$O^{\hat{\mathcal{Q}}}$ is then computed in the same way.
To this end, we deem the points whose confidence is greater than 
a threshold~$\theta_o$ to be within the overlap region.

\noindent {\bfseries Soft-Matching Module:} 
For the output features $\mathbf{H}^{\hat{\mathcal{P}}}_o$ and $\mathbf{H}^{\hat{\mathcal{Q}}}_o$ generated by the overlapping region detection module, we first normalize them to the unit hypersphere. Subsequently, we calculate the similarity matrix $\mathbf{S} \in \mathbb{R}^{|\hat{\mathcal{P}}| \times |\hat{\mathcal{Q}}|}$, where each element is defined as $s_{i,j} = \exp \left(-\left\|\mathbf{h}_i^{\hat{\mathcal{P}}}-\mathbf{h}_j^{\hat{\mathcal{Q}}}\right\|_2^2\right)$.
% For the output features $\mathbf{H}^{\hat{\mathcal{P}}}_o$, $\mathbf{H}^{\hat{\mathcal{Q}}}_o$ of the overlapping region detection module, we first normalize them to the unit hypersphere, and then calculate the similarity
% matrix: $\mathbf{S} \in \mathbb{R}^{|\hat{\mathcal{P}}|\times |\hat{\mathcal{Q}}|}$ with $s_{i,j}=\exp \left(-\left\|\mathbf{h}_i^{\hat{\mathcal{P}}}-\mathbf{h}_j^{\hat{\mathcal{Q}}}\right\|_2^2\right)$. 
% \SQ{to revise}
% Accordingly, we use separate normalization operations on different dimensions $0/1$ to allow one-to-many matches:
% \begin{equation}
% \bar{s}_{i, j}^{0/1}=\frac{s_{i, j}}{\sum_{k=1}^{|\hat{\mathcal{Q}}/\hat{P}|} s_{i, k/k,j}}, 
% \end{equation}
% Next, we extract purify correspondences by a threshold $\theta_{m}$:
% \begin{equation}\label{eq:purify}
%     \hat{\mathcal{C}}=\left\{(\hat{\mathbf{p}}_{x_{i}},\hat{\mathbf{q}}_{y_{i}})|(x_i,y_i)\in\|\bar{s}_{i, j}^{0/1}\ge \theta_{m}\|\right\}.
% \end{equation}
Accordingly, we apply softmax operation to similarity matrix $\mathbf{S}$ on two dimensions
separately to allow one-to-many matching. Next, we extract purify correspondences by a threshold $\theta_{m}$:

\begin{align}\label{eq:purify}
& \mathbf{S}_k = \operatorname{softmax}(\mathbf{S}(i,\cdot))_j, \\
& \hat{\mathcal{C}}_k = \left\{(\hat{\mathbf{p}}_i,\hat{\mathbf{q}}_j) |\mathbf{S}_k(i,j)\ge \theta_{m}\| \right\}
\end{align}
where $k \in 0,1$, $\mathbf{S}_0$ and $\mathbf{S}_1$ are the matching probability matrix
obtained by softmax operation along the first dimension and the zeroth dimension, $\hat{\mathcal{C}}_0$ and $\hat{\mathcal{C}}_1$ are the corresponding coarse-level correspondences proposals.Compared with the commonly used top-k selection strategy that needs to specify the number of matches, our strategy of using a tolerance can ensure that the number of selected correspondences is adaptive to the overlapping rate.

% where the hyperparameter~$\theta_{m}$ is set to ? in our experiments.
% In fact, the computation of~$\hat{\mathcal{C}}$ is inspired by lepard~\cite{Li2021LepardLP}.

% Instead of the Top-K selection strategy, inspired by lepard~\cite{Li2021LepardLP},
% Such an advantage is that compared with the former, it is difficult to select specific number of matches due to the impact of overlap rate of point cloud. For example, in low-overlap scenarios, larger k will often introduce more mismatches. So, 
It is important to acknowledge that while the previously mentioned strategy generates a larger number of potentially beneficial correspondences, it may lead to a low inlier ratio. To enhance this inlier ratio, we introduce a procedure where, for each superpoint in the source point cloud, we initially identify the most closely matched target superpoint based on $\mathbf{S}$, as well as the k-nearest neighbors of the target superpoint. Out of these $k+1$ correspondences, only those that satisfy the condition defined in Eq.~\ref{eq:purify} are retained. Similarly, for each superpoint in the target point cloud, this process is repeated until a pruned correspondence set $\hat{\mathcal{C}}_k$ is obtained. Finally, we further filter out mismatches outside predicted overlap regions $O^{\hat{\mathcal{P}}}$ and $O^{\hat{\mathcal{Q}}}$.

\subsection{Fine-level Matching}\label{sec.3.2}
% Since the vanilla attention~\cite{Vaswani2017AttentionIA} computing is expensive for high-resolution inputs.  To solve this problem, Linear Transformer~\cite{Katharopoulos2020TransformersAR} with linear complexity $O(N)$ is introduced as an alternative. In addition, we also incorporate positional information through relative position embedding operation~\cite{su2021roformer} to the inputs at each transformer layer to improve the position awareness of the model. In practically, we repeatedly exchange self- and cross-attention layer to update fine-level features for $\mathit{N}_{f}$ times to get reliable dense point matching.

\noindent {\bfseries Linear Transformer:} Linear Transformer~\cite{Wang2020LinformerSW} proposes to reduce the computation complexity by substituting the exponential kernel used in the original attention layer~\cite{Vaswani2017AttentionIA} with an alternative kernel function:
\begin{equation}
    \operatorname{sim}(Q,K) = \phi(Q) \cdot \phi(K)^T 
\end{equation}
where $\phi(\cdot)=elu(\cdot)+1$. Utilizing the associativity property of matrix products, the multiplication between $\phi(K)^T$ and $V$ can be carried out first. Sine $d_t \ll |\mathcal{P}|$, the computation cost is reduced to $O(|\mathcal{P}|)$.

Thanks to our overlap region detection module, we perform linear attention operations to improve feature discrimination only for points within the overlap region and not for all dense points. This reduces the impact of points in non-overlapping region on the one hand, and reduces the cost of calculation on the other hand. To be specific, we only focused on the points $\bar{\mathcal{P}}$ within patch $\{\mathcal{G}_{\mathbf{p}_{i}}^{\tilde{\mathcal{P}}}|\mathbf{p}_{i} \in O^{\hat{\mathcal{P}}}\}$ instead of all dense points $\tilde{\mathcal{P}}$, and the relevant features note as $\mathbf{F}^{\bar{\mathcal{P}}}$. We perform the same operation to get overlapping region points $\bar{\mathcal{Q}}$ and relevant features $\mathbf{F}^{\bar{\mathcal{Q}}}$.

Next, we adopt the linear transformer~\cite{Wang2020LinformerSW} to perform the self- and cross-attention to collect the global information through intra-and inter-relationship between features $\mathbf{F}^{\bar{\mathcal{P}}}$ and $\mathbf{F}^{\bar{\mathcal{Q}}}$. The self-attention layer updates its message by:
\begin{equation}
    \mathbf{Z}^{\bar{\mathcal{P}}} = \operatorname{LinearTransformer}(Q=K=V=\mathbf{F}^{\bar{\mathcal{P}}})
\end{equation}
and for $\mathbf{Z}^{\bar{\mathcal{Q}}},Q=K=V=\mathbf{F}^{\bar{\mathcal{Q}}}$. The cross-attention layer updates messages with information collected from the inter-relationship between two frame features:
\begin{equation}      
    \mathbf{Z}^{\bar{\mathcal{P}}} = \operatorname{LinearTransformer}(Q=\mathbf{F}^{\bar{\mathcal{P}}}, K=V=\mathbf{F}^{\bar{\mathcal{Q}}})
\end{equation}
and for  $\mathbf{Z}^{\bar{\mathcal{Q}}},Q=\mathbf{F}^{\bar{\mathcal{Q}}},K=V=\mathbf{F}^{\bar{\mathcal{P}}}$.

\noindent {\bfseries Relative Position Embedding:} Unlike the previous work, either choose to reduce the point cloud resolution~\cite{Li2021LepardLP, Yew2022REGTREP} to reduce the computing overhead of transformer, or only choose to improve the feature representation capability of superpoints~\cite{Li2021LepardLP, Qin2022GeometricTF}. Since we introduced linear transformer~\cite{Katharopoulos2020TransformersAR} to enhance the fine-level features, in order to improve the rotation invariance of the features, inspired by the work of lepard~\cite{Li2021LepardLP}, we also integrate rotation invariance information by adding rotation position embeddings~\cite{su2021roformer} to the inputs at each transformer layer, reducing the limitations on the rotation datasets. 

\noindent {\bfseries Hard-Matching Module:}  Through the above operation, we get a series of one-to-many superpoint correspondences located in overlapping regions, and the associated patches may have a low overlap rate, which will inevitably lead to a large number of dense point mismatches. Therefore, different from the soft matching strategy in Sec.~\ref{sec.3.1}, adopting a stricter matching strategy to suppress mismatches at fine-level is the key to obtain robust registration. Hence, we employ the point matching module~\cite{Qin2022GeometricTF}, which operates in conjunction with the optimal transmission strategy~\cite{Sinkhorn1967ConcerningNM}, to extract dense correspondences. The resultant correspondences set is denoted as $\mathcal{C}$. In addition, the  confidence score of $\mathcal{C}$ is denoted as $Z^ \mathcal{C}$.

\subsection{Feature-similar-based Efficient Registration}\label{sec.3.3}
 Robust pose estimators such as RANSAC~\cite{Fischler1981RandomSC} require a large number of iterations to guarantee accuracy, resulting in inefficiency. Considering the high inlier ratio of OAAFormer, we design an efficient estimator to achieve comparable performance while also greatly reducing the computational cost. This design is motivated by an important observation that a well-distributed set of correspondences where are more similar in the feature space is helpful for transform estimation.

\noindent {\bfseries Global sampling strategy:} In order to obtain the global sampling distribution, we perform the spectral matching technique~\cite{Leordeanu2005AST} to select reliable seeds, and the correspondences with local maximum confidence score within its neighborhood with radius $R$ are selected. The number of seed points $N_s$ is determined by the proportion of the whole correspondences $|\mathcal{C}|$. For each seed, we select its $k$-nearest neighbors in $Z^{\mathcal{C}}$ to expand into a consensus set. The total consensus sets can be noted as: $\mathcal{CS}\in \mathbb{R}^{N_s\times k}$

% we first calculate the Euclidean distance between each two points that in same point cloud per two correspondences.   
% $\{(x_i,x_j)|(i,j\in\mathcal{C}(x,\cdot))\}$ in $\tilde{\mathcal{P}}$ to form distance matrix: $\mathbf{D}^{\mathbf{E}}\in \mathbb{R}^{|\mathcal{C}|\times|\mathcal{C}|}$, where $d_{ij}^{e}= \left\|x_i-x_j\right\|_2 $. We then perform the spectral matching technique~\cite{Leordeanu2005AST} to select reliable seeds with confidence $\mathbf{Z}^{\mathcal{C}}$, and the correspondences with local maximum confidence score within its neighborhood with radius $R$ are selected. The number of seed points $N_s$ is determined by the proportion of the whole correspondences $|\mathcal{C}|$. For each seed, we select its $ck$-nearest neighbors in $\mathbf{Z}^{\mathcal{C}}$ to expand into a consensus set. The total consensus sets can be noted as: $\mathcal{CS}\in \mathbb{R}^{N_s\times ck}$

% $(\tilde{\mathbf{p}}_i,\tilde{\mathbf{q}}_j)$

\noindent {\bfseries Feature similarity compatibility:} We further analyzed the feature similarity of correspondences  in each consensus set. The intra-difference of each correspondence in one consensus set can be noted as : $\mathbf{D}^{\mathbf{F}}\in \mathbb{R}^{k \times 1}$,
% , where each element is defined as $d_k^f = \|\mathbf{f}_{x_{i}}-f_{y_{i}}\|^2_2$, 
and then we do a normalize as: $\mathbf{D}^{\mathbf{F}} = 1 - \mathbf{D}^{\mathbf{F}}/\operatorname{max}(\mathbf{D}^{\mathbf{F}})$. Accordingly, we use sigmoid operation to expand the inter-difference of correspondences as:
\begin{equation}
    \mathbf{D}^{\mathbf{F}}=\operatorname{sigmoid}((\mathbf{D}^{\mathbf{F}} - \operatorname{mean}(\mathbf{D}^{\mathbf{F}})\cdot \sigma_s)
\end{equation}
where $\sigma_s$ is parameter to control the sensitivity to difference of feature. In meanwhile, $\mathbf{D}^{\mathbf{F}}$ can be regarded as a feature similarity score. The more similar the correspondence feature is, the closer it is to 1, otherwise the closer it is to 0. Then we calculate the compatibility matrix of this consensus set and denoted it as $\mathbf{CM}\in \mathbb{R}^{k\times k}$
% . It's obvious that $\mathbf{CM}$ can be expressed as the minimum value of two correspondences scores.
, while each item of $\mathbf{CM}$ can be expressed as the minimum value of two correspondences
scores.

% \begin{equation}
%     cm_{i,j}=\operatorname{min}(d_i^f,d_j^f)
% \end{equation}

\noindent {\bfseries Hypothesis Selection:} The association of each correspondence with the leading eigen-vector is adopted as the weight for this correspondence and can be solved by power iteration algorithm~\cite{MisesPraktischeVD}. and then we use the weighted SVD~\cite{Arun1987LeastSquaresFO} on the consensus set to generate an estimation $(\mathbf{R}_i,\mathbf{t}_i)$ for each seed. Finally, we choose the transformation that allows the most correspondences in $\mathcal{C}$:
\begin{equation}
\mathbf{R}, \mathbf{t}=\max _{\mathbf{R}_i, \mathbf{t}_i} 
\begin{matrix}
\sum_{\left(\tilde{\mathbf{p}}_{j}, \tilde{\mathbf{q}}_{j}\right)\in\mathcal{C}}\llbracket\left\|\mathbf{R}_i \cdot \tilde{\mathbf{p}}_{j}+\mathbf{t}_i-\tilde{\mathbf{q}}_{j}\right\|_2^2<\tau_a \rrbracket
\end{matrix}
\end{equation}
where $\llbracket\cdot\rrbracket$ is the Iverson bracket. $\tau_a$ is the acceptance radius.

\subsection{Loss function:}\label{sec.3.5}
The final loss consists of the coarse-/fine-level loss and the overlap loss: $\mathcal{L} = \mathcal{L}_c + \mathcal{L}_f + 0.5 \ast \mathcal{L}_o$. As with geotransformer~\cite{Qin2022GeometricTF}, we use overlap-aware circle loss~\cite{Qin2022GeometricTF} and negative log-likelihood loss~\cite{Wang2019SuperGLUEAS} for coarse and fine level features, respectively. This also benefits us in allowing features to be closer between superpoints/patches with higher overlap ratios in coarse-level matching, rather than strictly limiting one-to-one matching. At the fine-level, stricter supervise can also help eliminate mismatches. Here, the overlap region estimation is regarded as a binary classification task, and the overlap loss $\mathcal{L}_o=\left(\mathcal{L}_o^{\hat{\mathcal{P}}}+\mathcal{L}_o^{\hat{\mathcal{Q}}}\right)/2$ is defined as:
\begin{equation}
\mathcal{L}_o^{\hat{\mathcal{P}}}=\frac{1}{|\hat{\mathcal{P}}|} \sum_{i=1}^{|\hat{\mathcal{P}}|} \bar{o}_{\hat{\mathbf{p}}_i} \log \left(o_{\hat{\mathbf{p}}_i}\right)+\left(1-\bar{o}_{\hat{\mathbf{p}}_i}\right) \log \left(1-o_{\hat{\mathbf{p}}_i}\right)
\end{equation}
The ground truth label $\bar{o}_{\hat{\mathbf{p}}_i}$ of superpoint $\hat{\mathbf{p}}_i$ is defined according whether it is in the ground-truth coarse matches set $\mathcal{A}$:
\begin{equation}
\bar{o}_{\hat{\mathbf{p}}_i}= \begin{cases}1, & \text{if} \ i \in \mathcal{A}(x, \cdot) \\ 0, & \text { otherwise }\end{cases}
\end{equation}
The reverse loss $\mathcal{L}_o^{\hat{\mathcal{Q}}}$ and  ground truth label $\bar{o}_{\hat{\mathbf{p}}_i}$ are computed in the same way.

\section{Experiments} In this section, we evaluate OAAFormer on indoor 3DMatch/3DLoMatch benchmarks (Sec~\ref{sec.4.1}), the outdoor KITTI odometry benchmark (Sec~\ref{sec.4.2}), and synthetic ModelNet/ModelLoNet benchmarks (Sec~\ref{sec.4.3}). For the coarse-level matching module, we repeatedly alternate between the geometric self-attention module~\cite{Qin2022GeometricTF} and the vanilla cross-attention module~\cite{Vaswani2017AttentionIA} by setting $N_c = 3$ and then pass through the overlap region detection module. Regarding the threshold $\theta_m$, we observed that $\theta_m = 0.05$ is safe to limit the number of superpoint matches to be within the range of $[256, 512]$. For k nearest neighbors, we find that $k=3$ achieves the best results. For fine-level matching, we also interleave the linear self-/cross-attention module by setting $N_f = 3$ to enhance feature discrimination. For the proposed efficient pose estimator, $\sigma_s=10$ is employed to augment the distinctiveness of correspondences, with $N_s$ being configured at 30\% of the total correspondences, and the parameter "$k$" is set to 20 for the establishment of the consensus set.

% In this section, 
% We evaluate OAAFormer on indoor 3DMatch/3DLoMatch benchmarks (Sec~\ref{sec.4.1}), outdoor KITTI odometry benchmark (Sec~\ref{sec.4.2}) and synthetic ModelNet/ModelLoNet benchmarks (Sec~\ref{sec.4.3}). For coarse-level matching module, we repeatedly interchange the geometric self-attention module~\cite{Qin2022GeometricTF} and the vanilla cross-attention module~\cite{Vaswani2017AttentionIA} by setting $N_c = 3$ and then pass through the overlapp region detection module. 
% In the setting of The threshold $\theta_m$,
% we observe that 
%  $\theta_m = 0.05$ is sate to limit the number of superpoint matches into $[256, 512]$. 
% For fine-level matching, we also interleave the linear self-/cross-attention module by setting $N_f = 3$ to enhance feature discrimination.

\subsection{Indoor Benchmark: 3DMatch}\label{sec.4.1}
\noindent {\bfseries Dataset. } 
3DMatch~\cite{Zeng20163DMatchLL} is a collection of 62 scenes, of which we employ 46 scenes for training, 8 for validation, and 8 for testing. We utilize the training data preprocessed by~\cite{Huang2020PREDATORRO} and conduct evaluations on both the 3DMatch and 3DLoMatch benchmarks. The former features a 30\% overlap, while the latter exhibits low overlap in the range of 10\% to 30\%. To assess robustness to arbitrary rotations, we follow~\cite{Wang2021YouOH} to create rotated benchmarks, where full-range rotations are independently applied to the two frames of each point cloud pair.
% 3DMatch~\cite{Zeng20163DMatchLL} is a collection of 62 scenes, from which we use 46 scenes for training, 8 for validation and 8 for testing. We use the training data preprocessed by~\cite{Huang2020PREDATORRO} and evaluate on both 3DMatch and 3DLoMatch benchmarks. The former is has 30\% overlap, and the latter is low overlap of 10\%$\sim$30\%. To evaluate robustness to arbitrary rotations, we follow~\cite{Wang2021YouOH} for creating the rotated benchmarks, where full-range rotations are individually added to the two frames of each point cloud pair.

\noindent {\bfseries Metrics. } 
We follow~\cite{Huang2020PREDATORRO,Yu2021CoFiNetRC,Qin2022GeometricTF} to employ three metrics for evaluation: (1) {\em Inlier Ratio} ({\bfseries IR}), which computes the ratio of putative correspondences with a residual distance smaller than a threshold ({\em i.e.}, 0.1m) under the ground-truth transformation; (2) {\em Feature 
 Matching Recall} ({\bfseries FMR}), which calculates the fraction of point cloud pairs with an {\bfseries IR} exceeding a threshold ({\em i.e.}, 5\%); and (3) {\em Registration Recall} ({\bfseries RR}) , which quantifies the fraction of point cloud pairs that are accurately registered ({\em i.e.}, with a root mean square error, {\bfseries RMSE} \textless 0.2m).

% We follow~\cite{Huang2020PREDATORRO,Yu2021CoFiNetRC,Qin2022GeometricTF} to use three metrics for evaluation: (1) {\em Inlier Ratio} ({\bfseries IR}) that computes the ratio of putative correspondences whose residual distance is smaller than a threshold ({\em i.e.}, 0.1m) under the ground-truth transformation; (2) {\em Feature 
%  Matching Recall} ({\bfseries FMR}) that calculates the fraction of point cloud pairs whose {\bfseries IR} is larger than a threshold ({\em i.e.}, 5\%); (3) {\em Registration Recall} ({\bfseries RR}) that counts the fraction of point cloud pairs that are correctly registered ({\em i.e.}, with {\bfseries RMSE} \textless 0.2m).

\begin{table}
\caption{Evaluation results on 3DMatch and 3DLoMatch.}
\vspace{2pt}
\setlength{\tabcolsep}{10.5pt}
\scriptsize
\centering
\begin{tabular}{l|cc|cc} 
\toprule
                 & \multicolumn{2}{c|}{3DMatch}          & \multicolumn{2}{c}{3DLoMatch}  \\
\# Samples=5,000 & Origin        & Rotated               & Origin        & Rotated        \\ 
\midrule
\multicolumn{5}{c}{\textit{ Feature Matching Recall} (\%)~}                               \\ 
\midrule
SpinNet~\cite{Ao2020SpinNetLA} & 97.4          & 97.4                  & 75.5          & 75.2           \\
Predator~\cite{Huang2020PREDATORRO} & 96.6          & 96.2                  & 78.6          & 73.7           \\
CoFiNet~\cite{Yu2021CoFiNetRC} & 98.1  & 97.4                  & 83.1          & 78.6           \\
YOHO~\cite{Wang2021YouOH} & \underline{98.2} & 97.8         & 79.4          & 77.8           \\
RIGA~\cite{Yu2022RIGARA} & 97.9          & \underline{98.2}                  & 85.1          & 84.5           \\
Lepard~\cite{Li2021LepardLP} & 98.0          & 97.4                  & 83.1          & 79.5           \\
GeoTrans~\cite{Qin2022GeometricTF}& 97.9       & 97.8         & \underline{88.3}  & \underline{85.8}   \\
OAAFormer(Ours)   & \textbf{98.6}  & \textbf{98.2}         & \textbf{89.8} & \textbf{89.5}  \\ 
\midrule
\multicolumn{5}{c}{\textit{ Inlier Ratio} (\%)}                                           \\ 
\midrule
SpinNet~\cite{Ao2020SpinNetLA}& 48.5          & 48.7                  & 25.7          & 25.7           \\
Predator~\cite{Huang2020PREDATORRO}& 58.0          & 52.8                  & 26.7          & 22.4           \\
CoFiNet~\cite{Yu2021CoFiNetRC}& 49.8          & 46.8                  & 24.4          & 21.5           \\
YOHO~\cite{Wang2021YouOH}& 64.4          & 64.1                  & 25.9          & 23.2           \\
RIGA~\cite{Yu2022RIGARA}& 68.4          & \underline{68.5}          & 32.1          & 32.1           \\
Lepard~\cite{Li2021LepardLP} & 58.6          & 53.7                  & 28.4          & 24.4           \\
GeoTrans~\cite{Qin2022GeometricTF} & \underline{71.9}  & 68.2                  & \underline{43.5}  & \underline{40.0}   \\
OAAFormer(Ours)   & \textbf{82.9} & \textbf{79.6}         & \textbf{50.1} & \textbf{48.2}  \\ 
\midrule
\multicolumn{5}{c}{\textit{ Registration Recall}~(\%)}                                    \\ 
\midrule
SpinNet~\cite{Ao2020SpinNetLA}& 88.8          & \underline{93.2}         & 58.2          & 61.8           \\
Predator~\cite{Huang2020PREDATORRO}& 89.0          & 92.0                  & 59.8          & 58.6           \\
CoFiNet~\cite{Yu2021CoFiNetRC}& 89.3          & 92.0                  & 67.5          & 62.5           \\
YOHO~\cite{Wang2021YouOH}& 90.8          & 92.5                  & 65.2          & 66.8           \\
RIGA~\cite{Yu2022RIGARA}& 89.3          & 93.0                  & 65.1          & 66.9           \\
Lepard~\cite{Li2021LepardLP}& 91.7 & 84.9                  & 62.5          & 49.0           \\
GeoTrans~\cite{Qin2022GeometricTF}& \underline{92.0}          & 92.0                  & \underline{75.0} & \underline{71.8}   \\
OAAFormer(Ours)   & \textbf{94.2}  & \textbf{93.8}   & \textbf{77.2}  & \textbf{76.0}  \\
\bottomrule
\end{tabular}
\label{tab.1}
% \vspace{-10pt}
\end{table}

\begin{table}
\caption{Evaluation results on 3DMatch and 3DLoMatch with a varying number of correspondences.}
\label{table:3DMatch}
\vspace{2pt}
\setlength{\tabcolsep}{2.2pt}
\scriptsize
\centering
\begin{tabular}{l|ccccc|ccccc}
\toprule
 & \multicolumn{5}{c|}{3DMatch} & \multicolumn{5}{c}{3DLoMatch} \\
\# Samples & 5000 & 2500 & 1000 & 500 & 250 & 5000 & 2500 & 1000 & 500 & 250 \\
\midrule
\multicolumn{11}{c}{\emph{Feature Matching Recall} (\%) $\uparrow$} \\
\midrule
PerfectMatch~\cite{Gojcic2018ThePM}  & 95.0 & 94.3 & 92.9 & 90.1 & 82.9 & 63.6 & 61.7 & 53.6 & 45.2 & 34.2 \\
FCGF~\cite{Choy2019FullyCG} & 97.4 & 97.3 & 97.0 & 96.7 & 96.6 & 76.6 & 75.4 & 74.2 & 71.7 & 67.3 \\
D3Feat~\cite{Bai2020D3FeatJL} & 95.6 & 95.4 & 94.5 & 94.1 & 93.1 & 67.3 & 66.7 & 67.0 & 66.7 & 66.5 \\
SpinNet~\cite{Ao2020SpinNetLA} & 97.6 & 97.2 & 96.8 & 95.5 & 94.3 & 75.3 & 74.9 & 72.5 & 70.0 & 63.6 \\
Predator~\cite{Huang2020PREDATORRO} & 96.6 & 96.6 & 96.5 & 96.3 & 96.5 & 78.6 & 77.4 & 76.3 & 75.7 & 75.3 \\
YOHO~\cite{Wang2021YouOH} &\underline{98.2} & 97.6 & 97.5 & 97.7 & 96.0 & 79.4 & 78.1 & 76.3 & 73.8 & 69.1 \\
CoFiNet~\cite{Yu2021CoFiNetRC} &98.1 &\underline{98.3} &\underline{98.1} & \underline{98.2} & \underline{98.3} & 83.1 & 83.5 & 83.3 & 83.1 & 82.6 \\
GeoTrans~\cite{Huang2020PREDATORRO} & 97.9 & 97.9 & 97.9 & 97.9  &97.6  &\underline{88.3} &\underline{88.6} &\underline{88.8} &\underline{88.6} &\underline{88.3} \\ % RGE with Proj
OAAFormer (\emph{ours}) &\textbf{98.6} &\textbf{98.6} &\textbf{98.5} &\textbf{98.5} &\textbf{98.2} &\textbf{89.8} &\textbf{89.9} &\textbf{90.1} &\textbf{90.1} &\textbf{89.9} \\ % RGE with Proj
\midrule
\multicolumn{11}{c}{\emph{Inlier Ratio} (\%) $\uparrow$} \\
\midrule
PerfectMatch~\cite{Gojcic2018ThePM} & 36.0 & 32.5 & 26.4 & 21.5 & 16.4 & 11.4 & 10.1 & 8.0 & 6.4 & 4.8 \\
FCGF~\cite{Choy2019FullyCG} & 56.8 & 54.1 & 48.7 & 42.5 & 34.1 & 21.4 & 20.0 & 17.2 & 14.8 & 11.6 \\
D3Feat~\cite{Bai2020D3FeatJL} & 39.0 & 38.8 & 40.4 & 41.5 & 41.8 & 13.2 & 13.1 & 14.0 & 14.6 & 15.0 \\
SpinNet~\cite{Ao2020SpinNetLA} & 47.5 & 44.7 & 39.4 & 33.9 & 27.6 & 20.5 & 19.0 & 16.3 & 13.8 & 11.1 \\
Predator~\cite{Huang2020PREDATORRO} & 58.0 & 58.4 &57.1 &54.1 & 49.3 &26.7 &28.1 &28.3 &27.5 & 25.8 \\
YOHO~\cite{Wang2021YouOH} &64.4 &60.7 & 55.7 & 46.4 & 41.2 & 25.9 & 23.3 & 22.6 & 18.2 & 15.0 \\
CoFiNet~\cite{Yu2021CoFiNetRC} & 49.8 & 51.2 & 51.9 & 52.2 &52.2 & 24.4 & 25.9 & 26.7 & 26.8 &26.9 \\
GeoTrans~\cite{Qin2022GeometricTF}  &\underline{71.9} &\underline{75.2} &\underline{76.0} &\underline{82.2} &\underline{85.1} &\underline{43.5} &\underline{45.3} &\underline{46.2} &\underline{52.9} &\underline{57.7} \\
OAAFormer (\emph{ours}) & \textbf{82.9} &\textbf{83.1} &\textbf{83.3} &\textbf{85.5} &\textbf{86.1} &\textbf{50.1} &\textbf{52.4} &\textbf{55.6} &\textbf{58.6} &\textbf{60.1} \\
\midrule
\multicolumn{11}{c}{\emph{Registration Recall} (\%) $\uparrow$} \\
\midrule
PerfectMatch~\cite{Gojcic2018ThePM}  & 78.4 & 76.2 & 71.4 & 67.6 & 50.8 & 33.0 & 29.0 & 23.3 & 17.0 & 11.0 \\
FCGF~\cite{Choy2019FullyCG} & 85.1 & 84.7 & 83.3 & 81.6 & 71.4 & 40.1 & 41.7 & 38.2 & 35.4 & 26.8  \\
D3Feat~\cite{Bai2020D3FeatJL} & 81.6 & 84.5 & 83.4 & 82.4 & 77.9 & 37.2 & 42.7 & 46.9 & 43.8 & 39.1 \\
SpinNet~\cite{Ao2020SpinNetLA} & 88.6 & 86.6 & 85.5 & 83.5 & 70.2 & 59.8 & 54.9 & 48.3 & 39.8 & 26.8 \\
Predator~\cite{Huang2020PREDATORRO} & 89.0 & 89.9 & 90.6 & 88.5 & 86.6 & 59.8 & 61.2 & 62.4 & 60.8 & 58.1 \\
YOHO~\cite{Wang2021YouOH} & 90.8 & 90.3 & 89.1 & 88.6 & 84.5 & 65.2 & 65.5 & 63.2 & 56.5 & 48.0 \\
CoFiNet~\cite{Yu2021CoFiNetRC} & 89.3 & 88.9 & 88.4 & 87.4 & 87.0 & 67.5 & 66.2 & 64.2 & 63.1 & 61.0 \\
GeoTransr~\cite{Qin2022GeometricTF} & \underline{92.0} & \underline{91.8} &\underline{91.8} &\underline{91.4} &\underline{91.2} & \underline{75.0} & \underline{74.8} & \underline{74.2} &\underline{74.1} &\underline{73.5} \\
OAAFormer (\emph{ours}) & \textbf{94.2} & \textbf{94.2} &\textbf{93.8} & \textbf{93.2} & \textbf{93.0} & \textbf{77.2} & \textbf{77.2} &\textbf{77.0} &\textbf{76.8} &\textbf{76.4} \\
\bottomrule
\end{tabular}
% \vspace{-5pt}
\end{table}

\begin{table}
\centering
\caption{Evaluation results on Rotated 3DMatch and 3DLoMatch with a varying number of correspondences.}
\label{table:3DRMatch}
\vspace{2pt}
\setlength{\tabcolsep}{2.1pt}
\scriptsize
\centering
\begin{tabular}{l|ccccc|ccccc} 
\toprule
                         & \multicolumn{5}{c|}{3DMatch(Rotated)}                                                  & \multicolumn{5}{c}{3DLoMatch(Rotated)}                                                  \\
\# Samples               & 5000          & 2500          & 1000          & 500           & 250           & 5000          & 2500          & 1000          & 500           & 250            \\ 
\midrule
\multicolumn{11}{c}{\textit{Feature Matching Recall} (\%) $\uparrow$}                                                                                                                     \\ 
\midrule
SpinNet~\cite{Ao2020SpinNetLA}                 & 97.4          & 97.4          & 96.7          & 96.5          & 94.1          & 75.2          & 74.9          & 72.6          & 69.2          & 61.8           \\
Predator~\cite{Huang2020PREDATORRO}                & 96.2          & 96.2          & 96.6          & 96.0          & 96.0          & 73.7          & 74.2          & 75.0          & 74.8          & 73.5           \\
YOHO~\cite{Wang2021YouOH}   & \underline{97.8}    &  97.8   & 97.4          & 97.6          & 96.4          & 77.8          & 77.8          & 76.3          & 73.9          & 67.3           \\
CoFiNet~\cite{Yu2021CoFiNetRC}                 & 97.4          & 97.4          & 97.2          & 97.2          & \underline{97.3}          & 78.6          & 78.8          & 79.2          & 78.9          & 79.2           \\
GeoTrans~\cite{Qin2022GeometricTF}                & \underline{97.8} & \underline{97.9} & \underline{98.1} & \underline{97.7}  & \underline{97.3}  & \underline{85.8}  & \underline{85.7}  & \underline{86.5}  & \underline{86.6}  & \underline{86.1}   \\
OAAFormer (\textit{ours}) & \textbf{98.2} & \textbf{98.2}  & \textbf{98.2}  & \textbf{98.1} & \textbf{98.1} & \textbf{89.8} & \textbf{89.6} & \textbf{89.6} & \textbf{89.4} & \textbf{89.2}  \\ 
\midrule
\multicolumn{11}{c}{\textit{Inlier Ratio} (\%) $\uparrow$}                                                                                                                                \\ 
\midrule
SpinNet~\cite{Ao2020SpinNetLA}                  & 48.7          & 46.0          & 40.6          & 35.1          & 29.0          & 25.7          & 23.9          & 20.8          & 17.9          & 15.6           \\
Predator~\cite{Huang2020PREDATORRO}                & 52.8          & 53.4          & 52.5          & 50.0          & 45.6          & 22.4          & 23.5          & 23.0          & 23.2          & 21.6           \\
YOHO~\cite{Wang2021YouOH}                    & 64.1          & 60.4          & 53.5          & 46.3          & 36.9          & 23.2          & 23.2          & 19.2          & 15.7          & 12.1           \\
CoFiNet~\cite{Yu2021CoFiNetRC}                 & 46.8          & 48.2          & 49.0          & 49.3          & 49.3          & 21.5          & 22.8          & 23.6          & 23.8          & 23.8           \\
GeoTrans~\cite{Qin2022GeometricTF}                & \underline{68.2}  & \underline{72.5}  & \underline{73.3}  & \underline{79.5}  & \underline{82.3}  & \underline{40.0}  & \underline{40.3}  & \underline{42.7}  & \underline{49.5}  & \underline{54.1}   \\
OAAFormer (\textit{ours}) & \textbf{82.9} & \textbf{82.9} & \textbf{83.3} & \textbf{83.3} & \textbf{83.5} & \textbf{48.2} & \textbf{48.5} & \textbf{50.4} & \textbf{52.3} & \textbf{54.6}  \\ 
\midrule
\multicolumn{11}{c}{\textit{Registration Recall} (\%) $\uparrow$}                                                                                                                         \\ 
\midrule
SpinNet~\cite{Ao2020SpinNetLA}                 & \underline{93.2} & \underline{93.2} & 91.1          & 87.4          & 77.0          & 61.8          & 59.1          & 53.1          & 44.1          & 30.7           \\
Predator~\cite{Huang2020PREDATORRO}                 & 92.0          & 92.8        & 92.0          & \underline{92.2} & 89.5          & 58.6          & 59.5          & 60.4          & 58.6          & 55.8           \\
YOHO~\cite{Wang2021YouOH}    & 92.5          & 92.3          & \underline{92.4}          & 90.2          & 87.4          & 66.8          & 67.1          & 64.5          & 58.2          & 44.8           \\
CoFiNet~\cite{Yu2021CoFiNetRC}                 & 92.0          & 91.4          & 91.0          & 90.3          & 89.6          & 62.5          & 60.9          & 60.9          & 59.9          & 56.5           \\
GeoTransformer~\cite{Qin2022GeometricTF}              & 92.0          & 91.9          & 91.8  & 91.5          & \underline{91.4}  & \underline{71.8}  & \underline{72.0}  & \underline{72.0}  & \underline{71.6}  & \underline{70.9}   \\
OAAFormer (\textit{ours}) & \textbf{93.8}  & \textbf{93.8}  & \textbf{93.6} & \textbf{93.6} & \textbf{93.2} & \textbf{76.0} & \textbf{75.4} & \textbf{75.4} & \textbf{75.3} & \textbf{74.9}  \\
\bottomrule
\end{tabular}
% \vspace{-5pt}
\end{table}

% We first compare the correspondence results of OAAFormer with the recent state-of-the-art in Tab.~\ref{tab.1}. For {\bfseries FMR}, our method achieves at least 2\% improvement on 3DLoMatch(Rotated), which means that the probability of successful registration in low-overlap cases by using estimators can be greatly improved. For {\bfseries IR}, our method surpasses the baselines consistently by 10\%$\sim$11\% on 3DMatch and 6\%$\sim$8\% on 3DLoMatch which means that the quality of the extracted correspondences are more credible.

\noindent {\bfseries Correspondence results. } 
We begin by comparing the results of OAAFormer with the recent state-of-the-art in Tab.~\ref{tab.1}, and then proceed to analyze the impact of varying the number of correspondences in Tab.~\ref{table:3DMatch} and Tab.~\ref{table:3DRMatch}. Notably, our method excels in terms of {\bfseries FMR}, outperforming all baselines significantly, particularly in the case of 3DLoMatch. This implies a substantial increase in the likelihood of achieving correct registration with our robust pose estimator in low-overlap scenarios, where we consistently find more than 5\% inliers. Furthermore, for {\bfseries IR}, our approach exhibits even more substantial improvements, surpassing all benchmarks by over 10\% on 3DMatch and more than 7\% on 3DLoMatch. It is worth mentioning that our method maintains a stable performance even when the number of correspondences varies. Additionally, due to our incorporation of rotational invariance position information during fine-level matching, we perform admirably on the rotated datasets.

% We first compare the results of OAAFormer with the recent state-of-the-art in Tab.~\ref{tab.1}, and then further report the influence of a varying number of correspondence in Tab.~\ref{table:3DMatch} and Tab.~\ref{table:3DRMatch}. For {\bfseries FMR}, our method significantly surpass all baselines, especially on the 3DLoMatch, which means that the probability of correct registration with the robust pose estimator in low overlap scenarios is greatly increased. In other words, we can find more than 5\% inliers in most cases. For {\bfseries IR}, our approach achieved even more significant improvements, beating all benchmarks by more than 10\% on 3DMatch and more than 7\% on 3DLoMatch. One point worth mentioning is that our method also has a stable performance when the number of correspondences is varied.  In addition, since we embed rotational invariance position information when fine-level matching, we also perform well on the rotated datasets.

\begin{table}
\caption{Registration results w/o RANSAC on 3DMatch (3DM) and 3DLoMatch (3DLM). The time overhead for transformation estimation is also provided.}
\label{tab.2}
\vspace{2pt}
\setlength{\tabcolsep}{5pt}
\scriptsize
\centering
\begin{tabular}{l|l|c|cc|c} 
\toprule
\multirow{2}{*}{Model} & \multicolumn{1}{c|}{\multirow{2}{*}{Estimator}} & \multirow{2}{*}{\#Samples} & \multicolumn{2}{c|}{RR(\%)}                        & \multicolumn{1}{c}{Time(s)\textbf{ }}  \\
                       & \multicolumn{1}{c|}{}                           &                            & \multicolumn{2}{c|}{3DM 3DLM}                      & \multicolumn{1}{c}{Pose}               \\ 
\midrule
SpinNet~\cite{Ao2020SpinNetLA}                  & RANSAC-50k  & 5000        & 88.6 & 59.8                               & \multirow{5}{*}{2.344}                  \\
Predator~\cite{Huang2020PREDATORRO}             & RANSAC-50k  & 5000        & 89.0          & 59.8                               &                                         \\
CoFiNet~\cite{Yu2021CoFiNetRC}                  & RANSAC-50k  & 5000        & 89.3          & 67.5                               &                                         \\
GeoTrans~\cite{Qin2022GeometricTF}              & RANSAC-50k& 5000          & \underline{92.0}          & \underline{75.0}                       &                                         \\
OAAFormer(ours)                                  & RANSAC-50k & 5000         & \textbf{94.2}  & \multicolumn{1}{l|}{\textbf{77.2}} & \multicolumn{1}{l}{}                    \\ 
\midrule
SpinNet~\cite{Ao2020SpinNetLA}                  & weighted SVD   & 250                        & 34.0          & 2.5                                & \multirow{5}{*}{0.008}                  \\
Predator~\cite{Huang2020PREDATORRO}             & weighted SVD & 250                        & 50.0          & 6.4                                &                                         \\
CoFiNet~\cite{Yu2021CoFiNetRC}                  & weighted SVD& 250                        & 64.6          & 21.6                               &                                         \\
GeoTrans~\cite{Qin2022GeometricTF}              & weighted SVD & 250                        & \underline{86.5}  & \underline{59.9}                       &                                         \\
OAAFormer(ours)                                  & weighted SVD & 250                        & \textbf{88.4} & \textbf{62.1}                      & \multicolumn{1}{l}{}                    \\ 
\midrule
CoFiNet~\cite{Yu2021CoFiNetRC}                  & \multicolumn{1}{c|}{LGR}                        & 5000                       & 85.5          & 63.2                               & \multirow{3}{*}{0.019}                  \\
GeoTrans~\cite{Qin2022GeometricTF}              & \multicolumn{1}{c|}{LGR}                        & 5000                       & \underline{91.2}  & \underline{73.4}                       &                                         \\
OAAFormer(ours)                                  & \multicolumn{1}{c|}{LGR}                        & 5000                       & \textbf{93.2} & \textbf{76.2}                      &                                         \\ 
\midrule
CoFiNet~\cite{Yu2021CoFiNetRC}                  & \multicolumn{1}{c|}{FSR}                        & 5000                       & 85.8          & 64.2                               & \multirow{3}{*}{0.022}                  \\
GeoTrans~\cite{Qin2022GeometricTF}              & \multicolumn{1}{c|}{FSR}                        & 5000                       & \underline{91.5}  & \underline{73.8}                       &                                         \\
OAAFormer(ours)                                  & \multicolumn{1}{c|}{FSR}                        & 5000                       & \textbf{93.4} & \textbf{76.8}                      &                                         \\
\bottomrule
\end{tabular}
\vspace{-5pt}
\end{table}

\noindent {\bfseries Registration results.} 
As Tab.~\ref{tab.1} shows, the primary metric related to the ultimate objective of point cloud registration is $\mathbf{RR}$. For this metric, we compute the transformation using RANSAC~\cite{Fischler1981RandomSC} with 50K iterations. OAAFormer excels in terms of $\mathbf{RR}$, outperforming the competition with significant margins. Specifically, we achieve improvements of 2.2\% on both the standard and rotated datasets for 3DMatch and even more remarkable enhancements of 2.2\% and 4.2\% on 3DLoMatch.

Additionally, we report registration recall under different numbers of correspondences in Tab.~\ref{table:3DMatch} and Tab.~\ref{table:3DRMatch}. It's evident that our method's performance is remarkably stable, eliminating the need for extensive correspondence sampling, as seen in previous methods aimed at performance improvement.

We then compare $\mathbf{RR}$ using RANSAC-free estimators in Tab.~\ref{tab.2}. We begin with weighted SVD~\cite{Arun1987LeastSquaresFO} over correspondences to solve for the alignment transformation. Thanks to high values of $\mathbf{FMR}$ and $\mathbf{IR}$, OAAFormer achieves $\mathbf{RR}$ scores of 88.4\% and 62.1\% on 3DMatch and 3DLoMatch, respectively, while the results of the baseline methods deteriorate significantly. This can be explained by the fact that, on one hand, the coarse-to-fine mechanism constrains the correspondences to specific patches rather than the global domain. On the other hand, our model further narrows down the correspondences to the overlapping region and enhances the discriminative capabilities of fine-level features.

Subsequently, we employ the local-to-global registration module (LGR)~\cite{Qin2022GeometricTF} and (FSR) in Sec.~\ref{sec.3.3} separately to compute the transformation. In comparison with LGR, the FSR in our method maintains a similar time cost but significantly improves the sampling distribution, making it more effective for transformation estimation and yielding higher $\mathbf{RR}$ scores. This efficient estimator delivers performance on par with the robust pose estimator (RANSAC) but with significantly lower time costs, offering over 100 times acceleration.

\noindent {\bfseries Ablation studies. } 
To gain a more comprehensive understanding of the individual modules within our method, we conducted a series of ablation studies. Following the methodology outlined in~\cite{Qin2022GeometricTF}, we introduced the metric {\em Patch Inlier Ratio} ($\mathbf{PIR}$) to measure the fraction of patch matches with actual overlap. Additionally, we introduced another metric, {\em Patch Overlap Precision} ($\mathbf{POP}$), to assess the precision of patches within the actual overlap. It's worth noting that the metrics $\mathbf{FMR}$ and $\mathbf{IR}$ were reported with correspondences in the set $\mathcal{C}$, while RANSAC~\cite{Fischler1981RandomSC} was employed for the registration process.

To investigate the effectiveness of the overlap detection module, we compared it with the MLP-directly module in Tab.~\ref{tab.3}. Leveraging the attention mechanism, our module has the capability to model the global overlap position, allowing for better perception of the overlap region. With a well-designed re-weighted prediction module, we obtained more accurate detection results for the overlap region. As accurate overlap estimation is pivotal for eliminating mismatches, our proposed module outperforms alternatives across all metrics.

Moving forward, to explore the interactions between the soft-matching module (SMM), overlapping detection module (ODM), and linear transformer module (LTM), we conducted relevant ablation experiments in Tab.~\ref{tab.4}. When all modules were removed, OAAFormer reverted to Geotransformer~\cite{Qin2022GeometricTF} and served as the baseline. In general, when we replaced the strict matching mechanism of the original implementation with SMM, due to the introduction of a one-to-many matching paradigm, while introducing a prior for local-to-local matching, some mismatches were inevitably introduced, resulting in a decline in all metrics. The introduction of ODM and LTM, on the other hand, enhanced the accuracy of coarse- and fine-level matching, respectively, and outperformed the original implementation. When all three modules were introduced simultaneously, SMM mined more potential patch matches, ODM eliminated mismatches distributed outside the estimated overlapping regions, and LTM made the dense features of the overlapping region more discriminative, achieving the best performance.

In addition, we replaced the relative position embedding~\cite{su2021roformer} with the absolute position embedding~\cite{Vaswani2017AttentionIA} in the linear attention module and conducted relevant ablation experiments. As shown in Tab.~\ref{table:PE}, in the context of the rotated version benchmark within the 3DMatch/3DLoMatch dataset, it is evident that the inclusion of relative position embedding resulted in superior performance. This observation suggests that incorporating relative positional information not only assists the neural network in effectively modeling distant spatial relationships but also enhances the network's capacity to discriminate between features within regions that are otherwise similar. Furthermore, it contributes to the augmentation of feature rotation invariance, thereby strengthening the network's robustness in handling variations in rotational transformations.

\begin{figure}[t]
  \centering
  \includegraphics[width=\linewidth]{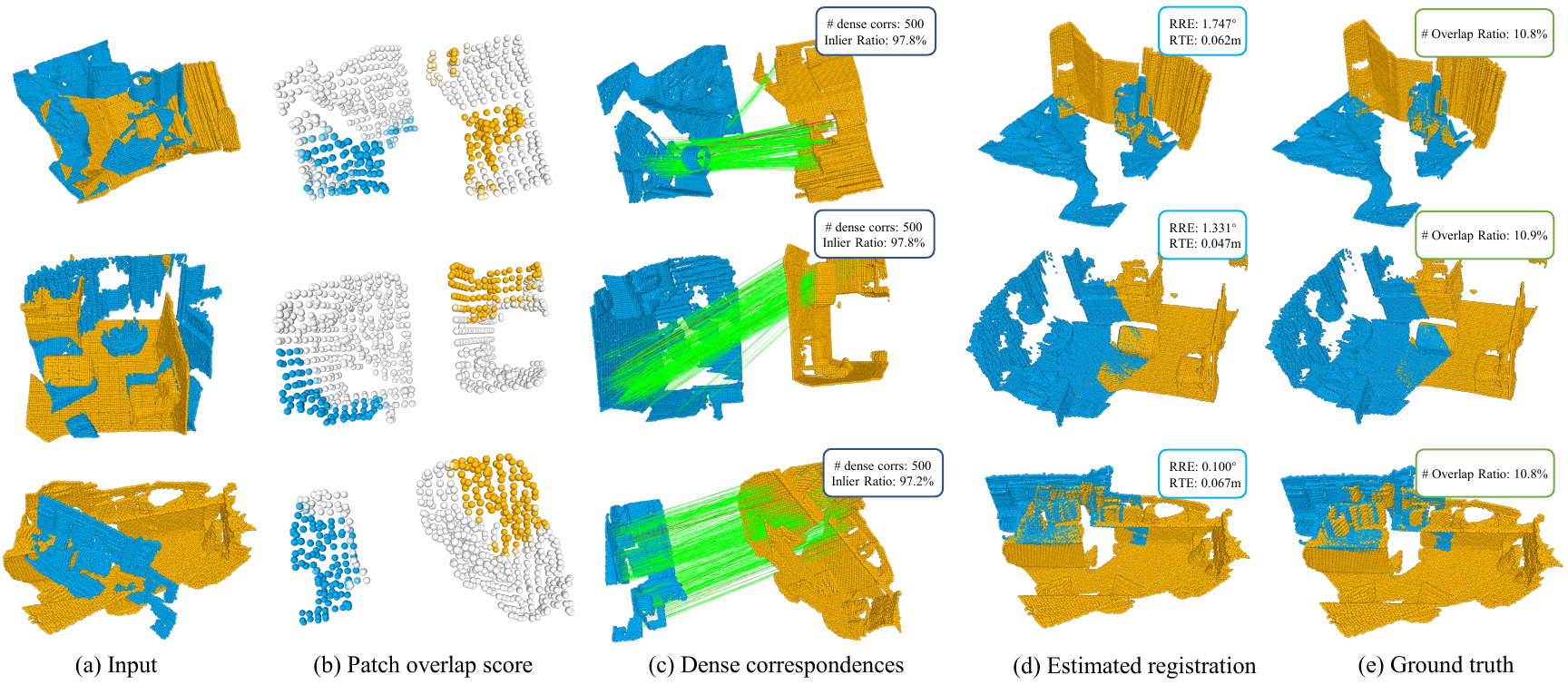}
  \vspace{-15pt}
  \caption{Qualitative results on 3DLoMatch. Column (b) shows the predicted overlap region, (c) represents the established correspondences, and (d) demonstrates the registration results. Green/red lines indicate inliers/outliers.}
  \label{fig.3}
\end{figure}

\begin{table}[t]
\caption{Ablation experiments of the overlapping region detection module.}
\label{tab.3}
\vspace{2pt}
\setlength{\tabcolsep}{0.8pt}
\scriptsize
\centering
\begin{tabular}{c|lcccc|lcccc} 
\toprule
\multirow{2}{*}{Model}       & \multicolumn{5}{c|}{3DMatch}     & \multicolumn{5}{c}{3DLoMatch}     \\
                             & POP   & PIR  & FMR  & IR   & RR   & OP   & PIR  & FMR  & IR   & RR    \\ 
\midrule
(a) MLP-directly               & 89.6 & 84.2 & 98.2 & 73.4 & 92.5 & 84.5 & 53.4 & 88.5 & 45.2 & 75.5  \\
(b) Overlap detection module & \textbf{93.5} & \textbf{85.6} & \textbf{98.6} & \textbf{82.9} & \textbf{94.2} & \textbf{88.1} & \textbf{54.2} & \textbf{89.8} & \textbf{50.1} & \textbf{77.2}  \\
\bottomrule
\end{tabular}
% \vspace{-10pt}
\end{table}

\begin{table}[t]
\caption{Ablation experiments of main modules.}
\label{tab.4}
% \vspace{-15pt}
\setlength{\tabcolsep}{3.5pt}
\scriptsize
\centering
\begin{tabular}{ccl|lccc|lccc} 
\toprule
\textbf{ } & \textbf{ } &                        & \multicolumn{4}{c|}{3DMatch}                                  & \multicolumn{4}{c}{3DLoMatch}                                  \\
SMM        & ODM        & LTM                    & PIR           & FMR           & IR            & RR            & PIR           & FMR           & IR            & RR             \\ 
\midrule
           &            &                        & \uline{86.1}  & 97.9          & 71.9          & 92.0          & \uline{54.9}  & 88.3          & 43.5          & 75.0           \\
\checkmark &            &                        & 82.7          & 97.4          & 68.0          & 91.3          & 46.4          & 86.1          & 38.1          & 73.5           \\
           & \checkmark &                        & \textbf{86.4} & 98.1          & 73.6          & 92.7          & \textbf{55.3} & 88.7          & 44.8          & 75.5           \\
           &            & \multicolumn{1}{c|}{\checkmark} & \uline{86.1}  & \uline{98.4}  & \uline{79.2}  & \uline{93.4}  & \uline{54.9}  & \uline{89.3}  & \uline{46.4}  & \uline{75.8}   \\
\checkmark & \checkmark & \multicolumn{1}{c|}{\checkmark} & 85.6          & \textbf{98.6} & \textbf{82.9} & \textbf{94.2} & 54.4          & \textbf{89.8} & \textbf{50.1} & \textbf{77.2}  \\
\bottomrule
\end{tabular}
\end{table}

\begin{table}
\centering
\caption{Ablation experiments of position embedding.}
\label{table:PE}
% \vspace{-15pt}
\setlength{\tabcolsep}{8pt}
\scriptsize
\centering
\begin{tabular}{l|ccc|ccc} 
\toprule
         & \multicolumn{3}{c|}{3DMatch(Rotated)}          & \multicolumn{3}{c}{3DLoMatch(Rotated)}                             \\
         & FMR  & IR  &RR & FMR  & IR  & RR  \\ 
\midrule
Absolute~\cite{Vaswani2017AttentionIA} & \textbf{98.0} & 80.2 & 93.2                    & 88.4           & 43.8          & 75.2                     \\
Relative~\cite{su2021roformer} & \textbf{98.2} & \textbf{82.9} & \textbf{93.8}          & \textbf{89.8}   & \textbf{48.2}          & \textbf{76.0}   \\
\bottomrule
\end{tabular}
\end{table}

% \begin{tabular}{ccl|lccc|lccc} 
% \toprule
% \textbf{ }           & \textbf{ }   &    & \multicolumn{4}{c|}{3DMatch}  & \multicolumn{4}{c}{3DLoMatch}   \\
% SMM   & ODM & LTM                                   & PIR           & FMR               & IR                & RR            & PIR            & FMR           & IR                & RR      \\ 
% \midrule
% &    &                                              & 86.1          & 97.9              & 70.3              & 92.0          & 54.9           & 88.3          & 43.5              & 75.0    \\ 
% \checkmark      &      &                            & 82.7          & 97.8              & 68.0              & 91.3           & 45.3           & 88.2          & 38.1              & 73.5    \\
% \checkmark    & \checkmark        &                 & 84.6          & 97.8              & 68.2              & 91.8          & 48.4           & 88.2          & 39.4              & 74.1   \\
% & \checkmark  & \checkmark                          & 87.0          & 98.2              & 79.2              & 93.4          & 55.4           & 88.4          & 46.4              & 75.8     \\
%  &   & \checkmark                                   & 86.1          & 98.1              & \textbf{83.1}     & 93.4          & 54.9           & \textbf{89.1} &\textbf{50.4}      &\underline{76.0}         \\
% \checkmark    & \checkmark   & \checkmark           & 85.6          & 98.1              & \underline{82.9}  & \textbf{94.2} & 53.2           & \textbf{89.1} &\underline{50.1}   &\textbf{77.2}             \\
% \bottomrule
% \end{tabular}
% \vspace{-5pt}
% \end{table}

\begin{figure*}[t]
  \centering
  \includegraphics[width=\linewidth]{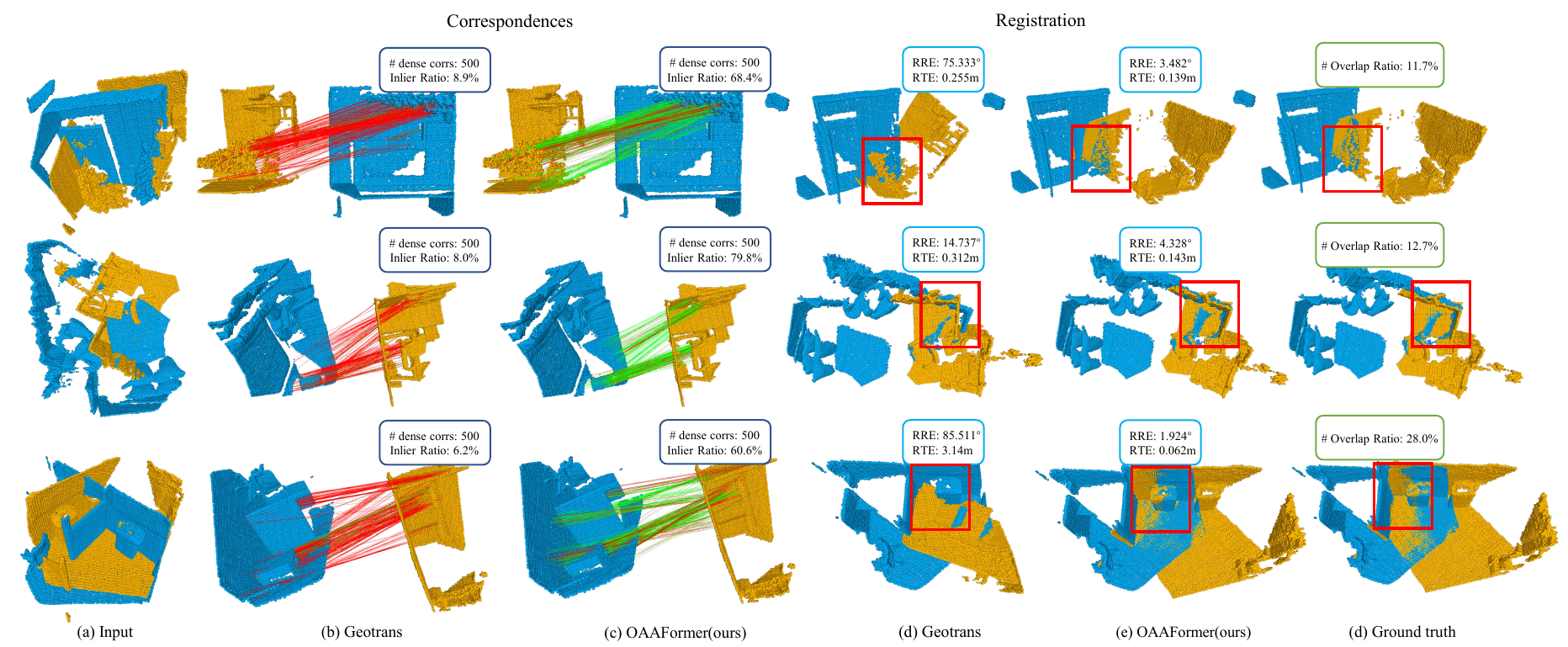}
  \vspace{-15pt}
  \caption{Qualitative results on 3DLoMatch.
  GeoTransformer~\cite{Qin2022GeometricTF} is used as the baseline. Columns (b) and (c) display the correspondences, while columns (d) and (e) illustrate the registration results. Green/red lines represent inliers/outliers.
  }
  \label{fig.4}
\end{figure*}

\noindent {\bfseries Qualitative results. } 
Fig.~\ref{fig.3} offers a visualization of the overlap region prediction in the coarse level and the dense correspondence results in the fine level. The overlapping region detection module excels in perceiving the global position, and the interaction module aids in determining whether superpoints are situated within the overlap region. Moreover, the Linear Transformer module with the relative position embedding strategy enhances the discriminative ability for dense correspondences, resulting in more reliable correspondences.

A gallery of registration and matching comparison results with state-of-the-art methods is shown in Fig.~\ref{fig.4}. It is evident that our method can establish more accurate correspondences across a broader spectrum of domains, yielding robust registration outcomes.

% Fig.~\ref{fig.3} provides a visualization of the overlap region prediction in coarse-level and the dense correspondences result in fine-level. The overlapping region detection module can better perceive the global position and the interaction module can help deduce whether superpoints are located in the overlap region. In addition, Linear transformer module with relative position embedding strategy helps to enhance the ability of distinguish for dense correspondences that more reliable correspondences are obtained.

% Fig.~\ref{fig.4} provides a gallery of the registration and matching comparison results with state-of-the-art methods. It shows that our method can establish more accurate correspondence in a wider range of domains and obtain more robust registration results simultaneously.

\subsection{Outdoor Benchmark: KITTI}\label{sec.4.2}
\noindent {\bfseries Dataset. } 
The KITTI odometry dataset~\cite{Geiger2012AreWR} comprises 11 sequences of LiDAR-scanned outdoor driving scenarios. For training, we adhere to the setup of~\cite{Pais20193DRegNetAD,Bai2020D3FeatJL}, utilizing sequences 0-5, while sequences 6-7 are reserved for validation, and sequences 8-10 are designated for testing. In line with the approach described in~\cite{Huang2020PREDATORRO}, we refine the ground-truth poses using ICP, and restrict the evaluation to point cloud pairs that are within a maximum distance of 10 meters.

% KITTI odometry~\cite{Geiger2012AreWR} contains 11 sequences of LiDAR-scanned outdoor driving scenarios. We follow~\cite{Pais20193DRegNetAD,Bai2020D3FeatJL} and use sequences 0-5 for training, 6-7 for validation, and 8-10 for testing. Similar to~\cite{Huang2020PREDATORRO}, the ground-truth poses are refined with ICP and we only use point cloud pairs that are at most 10m away for evaluation.

\noindent {\bfseries Metrics. }
We adhere to the evaluation metrics established by~\cite{Bai2020D3FeatJL,Huang2020PREDATORRO}, which include the following: 
(1) {\em Relative Rotation Error} ({\bfseries RRE}): This metric quantifies the geodesic distance between the estimated and ground-truth rotation matrices.
(2) {\em Relative Translation Error} ({\bfseries RTE}): It calculates the Euclidean distance between the estimated and ground-truth translation vectors.
(3) {\em Registration Recall} ({\bfseries RR}): This metric measures the fraction of point cloud pairs for which both {\bfseries RRE} and {\bfseries RTE} fall below specific thresholds, typically set as {\bfseries RRE}\textless $5^{\circ}$ and {\bfseries RTE}\textless 2 meters.
% We follow~\cite{Bai2020D3FeatJL,Huang2020PREDATORRO} to use three metrics for evaluation: (1) {\em Relative Rotation Error} ({\bfseries RRE}) that computes the geodesic distance between estimated and ground-truth rotation matrices. (2) {\em Relative Translation Error } ({\bfseries RTE}) that computes the Euclidean distance between estimated and ground-truth translation vectors, and (3) {\em Registration Recall } ({\bfseries RR}) that computes the fraction of point cloud pairs whose {\bfseries RRE} and {\bfseries RTE} are both below certain thresholds({\em i.e.}, {\bfseries RRE}\textless $5^{\circ}$ and {\bfseries RTE}\textless 2m).

\noindent {\bfseries Registration results. }
In Tab.~\ref{tab.5} (top), we compare OAAFormer with recent state-of-the-art methods, employing RANSAC as the pose estimator: D3Feat~\cite{Bai2020D3FeatJL}, SpinNet~\cite{Ao2020SpinNetLA}, Predator~\cite{Huang2020PREDATORRO}, CoFiNet~\cite{Yu2021CoFiNetRC}, and Geotransformer~\cite{Qin2022GeometricTF}. Our method performs comparably to these methods on {\bfseries RR} but outperforms the baseline by approximately 0.7 cm in terms of {\bfseries RTE} and 0.03° in {\bfseries RRE}. 
We also compare our method to three RANSAC-free methods in Tab.~\ref{tab.5} (bottom): FMR~\cite{Huang2020FeatureMetricRA}, DGR~\cite{Choy2020DeepGR}, HRegNet~\cite{Lu2021HRegNetAH}, and Geotransformer (with LGR)~\cite{Qin2022GeometricTF}. Our method outperforms all the baselines significantly. Furthermore, when using FSR as an estimator, our method surpasses all the RANSAC-based methods.

% In Tab.~\ref{tab.5}(top), we compare OAAFormer with the recent state-of-the-arts (RANSAC is used as pose estimator): D3Feat~\cite{Bai2020D3FeatJL}, SpinNet~\cite{Ao2020SpinNetLA}, Predator~\cite{Huang2020PREDATORRO}, CoFiNet~\cite{Yu2021CoFiNetRC} and Geotransformer~\cite{Qin2022GeometricTF}. Our method performs on-par with these method on {\bfseries RR}, but exceeded the baseline by about 0.7cm and 0.03° in terms of {\bfseries RTE} and {\bfseries RRE} respectively. We further compare to three RANSAC-free methods in Tab.~\ref{tab.5}(bottom): FMR~\cite{Huang2020FeatureMetricRA}, DGR~\cite{Choy2020DeepGR}, HRegNet~\cite{Lu2021HRegNetAH} and Geotransformer(w/ LGR)~\cite{Qin2022GeometricTF}. Our method outperforms all the baselines by large margin. In addition, our method with FSR as estimator beats all the RANSAC-based methods.

\begin{table}[h]
% \vspace{-5pt}
\caption{Registration results on KITTI odometry.}
\label{tab.5}
% \vspace{-15pt}
\scriptsize
\setlength{\tabcolsep}{4pt}
\centering
\begin{tabular}{l|ccc}
\toprule
Model & RTE(cm) & RRE($^{\circ}$) & RR(\%) \\
\midrule
3DFeat-Net~\cite{Yew20183DFeatNetWS} & 25.9 & \underline{0.25} & 96.0 \\
FCGF~\cite{Choy2019FullyCG} & 9.5 & 0.30 & 96.6 \\
D3Feat~\cite{Bai2020D3FeatJL} & 7.2 & 0.30 & \textbf{99.8} \\
SpinNet~\cite{Ao2020SpinNetLA} & 9.9 & 0.47 & 99.1 \\
Predator~\cite{Huang2020PREDATORRO} & 6.8 & 0.27 & \textbf{99.8} \\
CoFiNet~\cite{Yu2021CoFiNetRC} & 8.2 & 0.41 & \textbf{99.8} \\
GeoTrans~\cite{Qin2022GeometricTF} & \underline{7.4} & 0.27 & \textbf{99.8} \\
OAAFormer (\emph{ours}, RANSAC-\emph{50k}) & \textbf{6.6} & \textbf{0.24} & \textbf{99.8} \\
\midrule
FMR~\cite{Huang2020FeatureMetricRA} & $\sim$66 & 1.49 & 90.6 \\
DGR~\cite{Choy2020DeepGR}& $\sim$32 & 0.37 & 98.7 \\
HRegNet~\cite{Lu2021HRegNetAH}& $\sim$12 & 0.29 & 99.7 \\
GeoTrans (LGR)~\cite{Qin2022GeometricTF} & \underline{6.8} & \underline{0.24} & \textbf{99.8} \\
OAAFormer (\emph{ours}, FSR) & \textbf{6.0} & \textbf{0.21} & \textbf{99.8} \\
\bottomrule
\end{tabular}
\vspace{-15pt}
\label{table:kitti}
\end{table}

\subsection{Synthetic Benchmark: ModelNet}\label{sec.4.3}
\noindent {\bfseries Dataset. } 
ModelNet comprises 12,311 CAD models of synthetic objects spanning 40 distinct categories. We adhere to the practice of employing 5,112 samples for training, 1,202 samples for validation, and 1,266 samples for testing. Similar to~\cite{Huang2020PREDATORRO}, we conduct evaluations under two partial overlap scenarios: ModelNet, characterized by an average pairwise overlap of 73.5\%, and ModelLoNet, which exhibits a lower average overlap of 53.6\%.

% ModelNet contains 12,311 CAD models of synthetic objects from 40 different categories. We follow to use 5,112 samples for training, 1,202 samples for validation,
% and 1266 samples for testing. Following~\cite{Huang2020PREDATORRO}, we evaluate on two partial overlap settings: ModelNet which has 73.5\% pairwise overlap on average, and ModelLoNet which contains a lower 53.6\% average overlap.

\noindent {\bfseries Metrics. }
We adhere to the methodology outlined in~\cite{Huang2020PREDATORRO,Yew2022REGTREP} for performance evaluation, employing three key metrics: (1) {\bfseries RRE} (2) {\bfseries RTE} (with definitions consistent with those in Sec.~\ref{sec.4.2}), and (3) {\em Chamfer distance} ({\bfseries CD}), which quantifies the chamfer distance between two registered scans.

% We follow~\cite{Huang2020PREDATORRO,Yew2022REGTREP} to use three metrics for evaluation:(1)(2) {\bfseries RRE}, {\bfseries RTE} ({\em i.e.}, the definations are same as Sec.~\ref{sec.4.2}) (3) {\em Chamfer distance} ({\bfseries CD}) that computes the chamfer distance between two registered scans.

\noindent {\bfseries Registration results. } 
In Tab.~\ref{tab.6}, we conduct a comparative analysis of OAAFormer against state-of-the-art RANSAC-based methods and RANSAC-free methods. Notably, a few RANSAC-free methods are optimized primarily for ModelNet, and these models exhibit rapid performance deterioration in real-world scenarios. In contrast, OAAFormer demonstrates a substantial performance advantage over all baseline methods across all metrics, whether in the context of high overlap (ModelNet) or low overlap (ModelLoNet) scenarios.

% In Tab.~\ref{tab.6}, we compare OAAFormer to the state-of-the-art RANSAC-based methods and RANSAC-free methods. Several RANSAC-free methods are tuned specifically for ModelNet, and such models deteriorate rapidly in real-world scenarios, while OAAFormer significantly outperforms all baselines in all metrics under both high overlap (ModelNet) and low overlap (ModelLoNet) cases.

\begin{table}[h]
\caption{Registration results on ModelNet dataset}
\label{tab.6}
% \vspace{-15pt}
\scriptsize
\setlength{\tabcolsep}{2.3pt}
\centering
\begin{tabular}{l|ccc|ccc} 
\toprule
           & \multicolumn{3}{c|}{ModelNet}             & \multicolumn{3}{c}{ModelLoNet}            \\
Methods    & RRE            & RTE            & CD               & RRE            & RTE            & CD               \\ 
\midrule
Predator~\cite{Huang2020PREDATORRO}   &1.739          & 0.019          & 0.00089          & 5.235          & 0.132          & 0.0083           \\ 
OAAFormer(\emph{ours}, RANSAC-\emph{50k})   & \textbf{1.484} & \textbf{0.016} & \textbf{0.00081} & \textbf{4.143} & \textbf{0.091} & \textbf{0.0044}  \\
\midrule
PointNetLK~\cite{Li2021PointNetLKR} & 29.725         & 0.297          & 0.0235           & 48.567         & 0.507          & 0.0367           \\
OMNet~\cite{Xu2021OMNetLO}      & 2.947          & 0.032          & 0.0015           & 6.517          & 0.129          & 0.0074           \\
DCP-v2~\cite{Wang2019DeepCP}     & 11.975         & 0.171          & 0.0117           & 16.501         & 0.300          & 0.0268           \\
RPM-Net~\cite{Yew2020RPMNetRP}    & 1.712          & 0.018          & 0.00085          & 7.342          & 0.124          & 0.0050           \\
REGTR~\cite{Yew2022REGTREP}      &\underline{1.473}  &\underline{0.014} &\underline{0.00078}   &\underline{3.930}   &\underline{0.087}  &\underline{0.0037}\\ 
OAAFormer(\emph{ours}, FSR)   & \textbf{1.366} & \textbf{0.012} & \textbf{0.00074} & \textbf{3.884} & \textbf{0.074} & \textbf{0.0032}  \\
\bottomrule
\end{tabular}
\vspace{-20pt}
\end{table}

\section{Conclusions}

In this paper, we have enhanced the coarse-to-fine matching mechanism through a series of strategies. Key enhancements include (1) the development of a soft matching module to preserve valuable correspondences among superpoints, (2) the introduction of an overlapping region detection module for the elimination of mismatches and (3)  the incorporation of a region-wise attention module with linear complexity to bolster the discriminative capabilities of the extracted features. Furthermore, we propose a technique to accelerate the prediction process by carefully selecting limited but representative correspondences with high-confidence. Our method's effectiveness and robustness are validated through experiments conducted on three publicly available datasets.

{\small
\bibliographystyle{cvm}
\bibliography{cvmbib}
}

\end{document}